\documentclass[journal]{IEEEtran}
\ifCLASSINFOpdf
\else
\fi

\usepackage{threeparttable}
\usepackage{graphics} 
\usepackage{epsfig} 
\usepackage{amsmath} 

\newcommand\norm[1]{\left\lVert#1\right\rVert}
\usepackage{xcolor,colortbl}
\usepackage[caption=false]{subfig}
\definecolor{Gray}{gray}{0.94}
\usepackage{tikz}
\def\checkmark{\tikz\fill[scale=0.4](0,.35) -- (.25,0) -- (1,.7) -- (.25,.15) -- cycle;} 
\usepackage{multirow}

\begin{document}
%
\title{DeepTIO: A Deep Thermal-Inertial Odometry with Visual Hallucination}
%
%
%

\author{Muhamad Risqi U. Saputra, Pedro P. B. de Gusmao, Chris Xiaoxuan Lu, Yasin Almalioglu, \\ Stefano Rosa, Changhao Chen, Johan Wahlstr\"om, Wei Wang, Andrew Markham, and Niki Trigoni
\thanks{This work is supported by the National Institute of Standards and Technology (NIST) grant ``Pervasive, Accurate, and Reliable Location-Based Services for Emergency Responders'' (No. 70NANB17H185). M. R. U. Saputra was funded by Indonesia Endowment Fund for Education (LPDP).}
\thanks{The authors are with the Department of Computer Science, University of
Oxford, UK {\tt\small firstname.lastname@cs.ox.ac.uk}}
}

%
%

\markboth{Journal of \LaTeX\ Class Files}%
{Saputra \MakeLowercase{\textit{et al.}}: Bare Demo of IEEEtran.cls for IEEE Journals}
%



\maketitle

\begin{abstract}
Visual odometry shows excellent performance in a wide range of environments. However, in visually-denied scenarios (e.g. heavy smoke or darkness), pose estimates degrade or even fail. Thermal cameras are commonly used for perception and inspection when the environment has low visibility. However, their use in odometry estimation is hampered by the lack of robust visual features. In part, this is as a result of the sensor measuring the ambient temperature profile rather than scene appearance and geometry. To overcome this issue, we propose a Deep Neural Network model for thermal-inertial odometry (DeepTIO) by incorporating a visual hallucination network to provide the thermal network with complementary information. The hallucination network is taught to predict fake visual features from thermal images by using Huber loss. We also employ selective fusion to attentively fuse the features from three different modalities, i.e thermal, hallucination, and inertial features. Extensive experiments are performed in hand-held and mobile robot data in benign and smoke-filled environments, showing the efficacy of the proposed model.
\end{abstract}


%
\IEEEpeerreviewmaketitle

\section{Introduction}
%
%
%
%
\IEEEPARstart{C}{amera} pose estimation is a key enabler for a wide range of applications in robotics and computer vision. Primary examples include position tracking of mobile robots, autonomous vehicles, pedestrians, or mobile devices for augmented reality applications. Visual Odometry (VO) is the \emph{de facto} solution for estimating camera pose. Many VO techniques have been proposed, ranging from traditional feature-based approaches \cite{Nister2004f, Forster2014, Mur-Artal2015b} to the more recently developed Deep Neural Network (DNN) based approaches \cite{Wang2017, saputra19clvo, almalioglu2018ganvo, zhan2018unsupervised}. While VO is useful in a number of scenarios, its application is limited to those with sufficient illumination. For instance, VO systems fail in locating aerial robots in dim underground tunnels \cite{kanellakis2016evaluation} or tracking a firefighter in emergency response scenarios in presence of airborne particulates (e.g. smoke and soot). In contrast, thermal cameras are not affected by illumination conditions or airborne particulates, making them a viable sensing alternative to RGB cameras.

Although thermal cameras have been commonly used in visually-denied environments, their use cases are largely limited to perception and inspection \cite{peynot2009towards, quater2014light}. The main hindrance preventing their usage in odometry estimation is the lack of visual features (e.g. edges and textures) in the imaging system. Thermal cameras capture the radiation emitted from objects in the Long-Wave Infrared (LWIR) portion of the spectrum. These raw radiometric data are then converted to a temperature profile represented in a visible format (e.g. grayscale) to ease human interpretation \cite{yu2013camera}. As the camera captures the environmental temperature rather than the scene appearance and geometry, it is difficult to extract sufficient hand engineered features to accurately estimate pose. Moreover, even for the same scene, the extracted features are dependent on the temperature gradient. This issue is further compounded by the fact that every thermal camera is plagued with fixed-pattern noise and requires frequent re-calibration during operation through Non-Uniformity Correction (NUC) which periodically freezes the images for about half to one second \cite{averbuch2007scene} every 30-150~seconds.

\begin{figure}
    \centering
    \includegraphics[width=0.95\columnwidth]{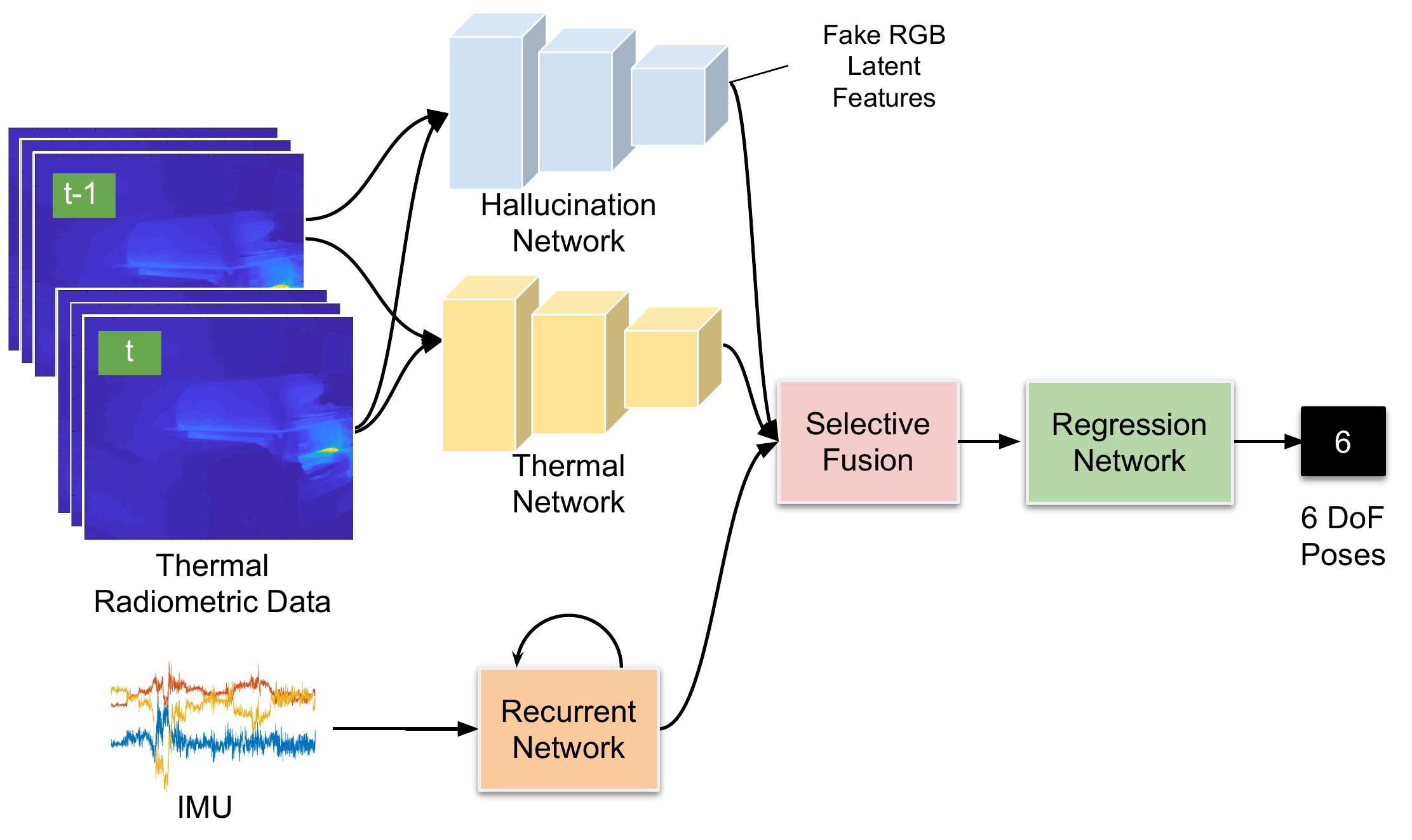}
    \caption{The architecture of DeepTIO at test time. DeepTIO not only extracts thermal features but also hallucinates visual features to provide additional information for accurate odometry.}
\label{fig:deeptio_arch_test}
\end{figure}

The last decade has witnessed a rapid development in the use of deep learning for automatically extracting salient features by directly learning a non-linear mapping function from data. We believe that, with sufficient training data, a DNN can also learn to infer 6 Degree-of-Freedom (DoF) poses from a sequence of thermal images. However, despite the DNN's ability to model this complexity, as stated before, thermal images are largely textureless and inherently lack sufficient features for accurate odometry estimation. 
Our novel intuition to alleviate this issue is to force our network to not only extract features from thermal images, but to additionally learn to hallucinate visual features similar to the ones extracted from a DNN-based VO, which have been proven to work well \cite{Wang2017, saputra19clvo, almalioglu2018ganvo, saputra2018visual}. Given sufficient training data, we hypothesize that hallucinating visual features is possible and can provide the thermal network with auxiliary information for accurate odometry estimation.

In this paper, we propose a DNN-based thermal-inertial odometry which is able to estimate accurate camera pose by not only extracting features from thermal images, but also hallucinating the visual features given thermal images as input. We also fuse the thermal image stream with Inertial Measurement Unit (IMU) data to improve pose estimation robustness due to its environment-agnostic characteristic. To this extent, we employ selective fusion \cite{chen2019selective} to adaptively fuse the different modalities conditioned on the input data. In summary, our key contributions are as follows:
\begin{itemize}
    \item We propose the first end-to-end trainable Deep Thermal-Inertial Odometry (DeepTIO) model.
    \item We present a novel deep neural odometry architecture incorporating a hallucination network.
    \item We present a new application of selective fusion with input from three feature channels, i.e. thermal, IMU, and hallucinated visual features.
    \item We perform extensive experiments and analysis in our self-collected hand-held and mobile robot dataset in benign and smoke-filled environments.
\end{itemize}

\section{RELATED WORK}

\subsection{Thermal Odometry}
Accurately estimating camera ego-motion from a thermal imaging system remains a challenging problem. Some efforts have been made towards thermal odometry systems, although these are limited to relatively short distances and yield sub-optimal performances compared to visible camera systems. Existing works either rely on sparse feature-based or direct-based approaches. Mouats et al. \cite{mouats2015thermal} employed a Fast-Hessian feature detector for UAV tracking using a stereo thermal camera. Khattak et al. \cite{khattak2019keyframe} developed a keyframe-based direct approach which minimizes radiometric error (raw thermal data) between consecutive frames. Borges and Vidas \cite{borges2016practical} designed a practical thermal odometry system with an automatic mechanism to determine when to perform NUC operation based on the current and the predicted poses. To improve robustness during NUC operation, most works incorporate thermal imaging with other modalities such as visual \cite{poujol2016visible, khattak2019visual} or inertial \cite{papachristos2018thermal, khattak2019keyframe}.

\subsection{DNN-based Odometry}

Due to the advancements of DNN, learning based odometry is recently gaining more favor. Wang et al.~\cite{Wang2017} started this trend by introducing an end-to-end trainable deep visual odometry method (DeepVO) by composing the feature extraction capabilities of Convolutional Neural Networks (CNN) and the ability to model long-term camera pose dependencies using a Recurrent Neural Network (RNN). This was then followed by other improvements such as enforcing consistency among multiple poses \cite{saputra19clvo} or introducing additional learning signals by performing multi-task learning such as global pose localization \cite{valada2018deep} or semantic segmentation \cite{radwan2018vlocnet}. Other works improved the robustness of VO by fusing visual and inertial streams \cite{clark2017vinet} and performing selective fusion between visual and inertial features \cite{chen2019selective}. In parallel to these supervised approaches, many self-supervised DNN-based VO approaches were also developed by leveraging the view reconstruction paradigm, started by Zhou et al. \cite{Zhou2017}, followed by adding stereo information \cite{li2018undeepvo, zhan2018unsupervised, yang2018deep} or generative networks \cite{almalioglu2018ganvo, feng2019sganvo}. While many works exist for DNN-based odometry, to the best of our knowledge none of them uses thermal camera as input.

\subsection{Learning with Side Information}

Related to our work is the concept of learning with side information. Hoffman et al. \cite{hoffman2016learning} introduced this concept by incorporating a depth hallucination network to increase the accuracy of object detection in RGB images. This concept was then adopted in other applications such as learning hand articulations \cite{choi2017learning} or face recognition \cite{lezama2017not}. Our work introduces this concept to odometry regression and trains the whole network with the non-trivial Huber loss. We are the first to hallucinate visual features from thermal images for odometry regression.

\section{NETWORK ARCHITECTURE}

In this section we describe our proposed DeepTIO model for estimating thermal-inertial odometry. Fig. \ref{fig:deeptio_arch_test} illustrates the general architecture of DeepTIO model at inference time. 
It is composed by a feature encoder, a selective fusion module, and a pose regressor. 
The feature encoder extracts salient features from each modality. We use a CNN for encoding thermal data and hallucinating visual features from thermal images. To extract features from the IMU data stream we employ a RNN, as RNN works better to model temporal dependencies of time-series data \cite{connor1994recurrent}. The feature vectors generated from the IMU, thermal, and hallucination encoder networks are input into the selective fusion module, attentively selecting certain features that are necessary for pose regression. The reweighted features are further feed into pose regression module to infer 6-DoF relative camera poses. 
The details of each module are described as below.

\subsection{Feature Encoder}
\label{sec:feature_encoder}

Given a pair of consecutive thermal images $\textbf{x}_{T} \in {\rm I\!R}^{2 \times ( w \times h \times c)}$, the purpose of the thermal encoder network is to extract geometrically meaningful features for movement estimation (e.g. optical flow captured between moving edges). To this end, both thermal encoder $\Psi_{T}$ and hallucination encoder $\Psi_{H}$ are implemented and pre-initialized with FlowNetSimple structure \cite{Dosovitskiy2016}. As the observed temperature profile (in grayscale) fluctuates when the camera captures hotter objects, we directly use the 16 bit raw radiometric data to obtain more stable inputs. Since raw radiometric data are only represented by one channel, we duplicate it into three channels for feeding into the FlowNet structure. We use the last output activation from both $\Psi_{T}$ and $\Psi_{H}$ as our thermal $\textbf{a}_{T}$ and visual hallucinated $\textbf{a}_{H}$ features
\begin{align}
  \label{eq:thermal_features}
  \textbf{a}_{T} = \Psi_{T} ( \textbf{x}_{T} ), \quad \textbf{a}_{H} = \Psi_{H} ( \textbf{x}_{T} ) .
\end{align}
We employ a single LSTM layer with 256 hidden states as IMU encoder $\Psi_{I}$. 
The 6-dimensional inertial data with a sequence of 20 frames $\textbf{x}_{I} \in {\rm I\!R}^{6 \times 20}$ are fed into IMU encoder $\Psi_{I}$ to produce IMU features
\begin{align}
  \label{eq:imu_features}
  \textbf{a}_{I} = \Psi_{I} ( \textbf{x}_{I} ) .
\end{align}
To balance the number of features, we perform average pooling for $\textbf{a}_{T}$ and $\textbf{a}_{H}$, such that the final dimensions for all features are $\textbf{a}_{T} \in {\rm I\!R}^{2048}$, $\textbf{a}_{H} \in {\rm I\!R}^{2048}$, and $\textbf{a}_{I} \in {\rm I\!R}^{5120}$. 

\begin{figure}
    \centering
    \includegraphics[width=0.95\columnwidth]{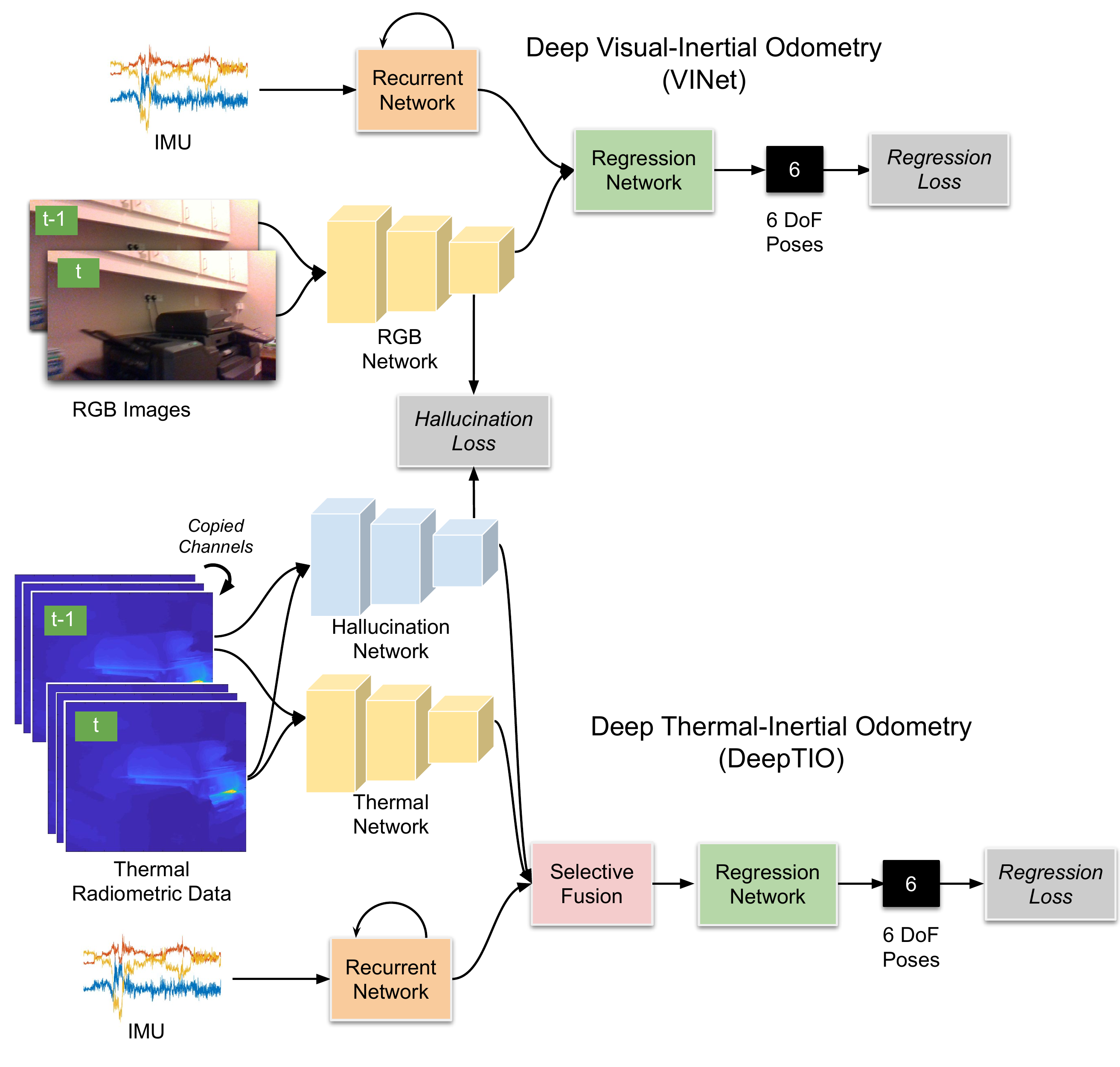}
    \caption{The architecture of DeepTIO at training time. Note how RGB images are used to guide the visual hallucination.}
\label{fig:deeptio_arch_train}
\end{figure}

\subsection{Selective Fusion}
\label{sec:selective_fusion}
In deep learning-based VIO, a standard way to fuse feature vectors coming from different modalities is by concatenation. However, a direct fusion of all feature modalities using concatenation results in sub-optimal performance, as not all features are useful and necessary \cite{chen2019selective}. The situation is even more exasperated by the intrinsic noise distribution of each modality. In our case, thermal data are plagued by fixed-pattern noise, while IMU data are affected by white random noise and sensor bias. On the other hand, the hallucination network might produce erroneous visual features. Moreover, in real applications there is high chance that different modalities, as well as the ground truth poses, will not be tightly synchronized. 

To this end, we employ selective fusion \cite{chen2019selective} to let the network automatically learn the best suitable feature combination given feature inputs. Specifically, a deterministic soft fusion is employed to attentively fuse features from three sources with compensation for possible misalignment between inputs and ground truth. The fusion module will learn to re-weight each feature by conditioning on all channels. The corresponding mask for thermal $\textbf{m}_{T}$, hallucination $\textbf{m}_{H}$, and inertial feature $\textbf{m}_{I}$ are learnt via:
\begin{align}
  \label{eq:mask_generation}
  \begin{split}
      \textbf{m}_{T} & = \sigma (\textbf{W}_{T}  [\textbf{a}_{T}; \textbf{a}_{H}; \textbf{a}_{I}] )  \\
  \textbf{m}_{H} & = \sigma (\textbf{W}_{H} [\textbf{a}_{T}; \textbf{a}_{H}; \textbf{a}_{I} ]) \\
  \textbf{m}_{I} & = \sigma (\textbf{W}_{I} [\textbf{a}_{T}; \textbf{a}_{H}; \textbf{a}_{I} ]),
  \end{split}
\end{align}
where $[\textbf{a}_{T}; \textbf{a}_{H}; \textbf{a}_{I}]$ denotes the concatenation of all channels features, $\sigma(x) = 1/( 1 + e^{-x})$ is the sigmoid function and $\textbf{W}_{T}$, $\textbf{W}_{H}$, and $\textbf{W}_{I}$ are the learnable weights for each feature modality. These masks are used to weight the relative importance of the features modalities by multiplying them via element-wise operation $\odot$ with their corresponding masks:
\begin{align}
  \label{eq:mask_generation}
  \textbf{a}_{fused} & = [ \textbf{a}_{T} \odot \textbf{m}_{T}; \textbf{a}_{H} \odot \textbf{m}_{H}; \textbf{a}_{I} \odot \textbf{m}_{I} ].
\end{align}
Finally, the merged features $\textbf{a}_{fused}$ are fed to the pose regressor network to estimate 6-DoF poses. 

\subsection{Pose Regressor}
The pose regressor consists of LSTM layers followed by two parallel Fully Connected (FC) layers that estimate relative translation and rotation respectively. We use an LSTM to model the long-term temporal dependencies of camera ego-motion as seen in \cite{Wang2017, saputra19clvo}. Each LSTM has 512 hidden states and takes the reweighted features $\textbf{a}_{fused}$ as input. The output latent vectors from the LSTMs are then fed into three parallel FC layers with 128, 64, 3 units respectively. We decouple the FC layers for translation and rotation as it has been shown to work better separately as in \cite{deeppco2019}. We also use a dropout \cite{srivastava2014dropout} rate of 0.25 between FC layers to help regularization.

\section{LEARNING MECHANISM}
This section introduces the mechanism to train hallucination network and learn odometry regression.

\subsection{Learning Visual Hallucination}
\label{sec:learn_hallucination}

The visual hallucination network $\Psi_{H}$ is intended to provide additional information along with the thermal encoder $\Psi_{T}$. Given original thermal images $\textbf{x}_{T}$ as an input, this module produces visual hallucination vectors $\textbf{a}_{H}$ that imitate the visual features $\textbf{a}_{V}$ from real RGB image input encoded by a visual encoder $\Psi_{V}$. 
In order to acquire pseudo ground truth of visual features, we employ a modified deep Visual-Inertial Odometry (VIO) model, i.e. VINet \cite{clark2017vinet}. 
The only difference is that we utilize FlowNetSimple as the feature extractor instead of FlowNetCorr \cite{Dosovitskiy2016} as used in the original VINet. This modification allows hallucination features $\textbf{a}_{H}$ and visual features $\textbf{a}_{V}$ to have same dimension, simplifying the training process.
After training VINet model, the weights $\textbf{W}_{V}$ in visual encoder $\Psi_{V}$ are frozen during the training of hallucination network, while the hallucination encoder $\Psi_{H}$'s weights $\textbf{W}_{H}$ are trainable.

Fig. \ref{fig:deeptio_arch_train} illustrates the architecture of our visual hallucination model in training process. We train the hallucination network by minimizing the discrepancy $\xi$ between the output activation from $\Psi_{H}$ and $\Psi_{V}$. Standard $\mathcal{L}_2$ norm is generally used for minimizing $\xi$ in benign cases \cite{hoffman2016learning, saputra2019distilling}. However, thermal camera requires periodic NUC calibration, during which time the same image will be output for between half to one second. NUC will force several identical thermal features to be matched with different visual features during network training. This process might produce an erroneous mapping between $\textbf{a}_{H}$ and $\textbf{a}_{V}$ and contaminate $\xi$ with outliers. Since the $\mathcal{L}_2$ loss is very sensitive to outliers, encountering some during training will impact gradient back-propagation as the outliers will dominate the loss, impacting convergence. To improve robustness against outliers, we instead propose to use the Huber Loss $\mathcal{H}$ \cite{huber1992robust} to minimize $\xi$. Then, our hallucination loss $\mathcal{L}_{\text{hallucinate}}$ is formally defined as follows:
\begin{align}
\begin{split}
    & \mathcal{L}_{\text{hallucinate}} = \frac{1}{n} \sum_{i=1}^{n} \mathcal{H}_{i} (\xi), \\
    & \text{with} \quad \xi = \Psi_{H}(\textbf{x}_{T};\textbf{W}_{H}) - \Psi_{V}(\textbf{x}_{V};\textbf{W}_{V}), \\
    & \text{and} \quad \mathcal{H} (\xi) = \begin{cases}
    \frac{1}{2}\norm{\xi}^{2} &  \text{for} \norm{\xi} \leq \delta, \\
    \delta ( \norm{\xi} - \frac{1}{2} \delta )              & \text{otherwise}
     \end{cases} 
\label{eq:hallucination_loss}
\end{split}
\end{align}
where $\delta$ is a threshold and $n$ is the batch size during training. By using Huber loss, $\xi$ larger than $\delta$ will have a linear effect instead of quadratic, making it less sensitive to outliers. Loss values below $\delta$ will still be minimized using quadratic loss to enable fast convergence. During training, we use $\delta = 1.0$.

\subsection{Learning Odometry Regression}
\label{sec:learn_pose}

We train the network to estimate odometry by minimizing the loss between the predicted pose and the ground truth pose. This task is essentially learning a mapping function from the input to the output $\{ (\textbf{x}_T;\textbf{x}_I)_{1:N} \} \rightarrow \{ ( {\rm I\!R}^{6} )_{1:N} \}$ where $N$ is the whole training data. The pose regressor network, together with all other networks except the hallucination part, are trained using the following regression loss
\begin{align}
\label{eq:reg_loss}
     \mathcal{L}_{regress} & = \frac{1}{n} \sum_{i=1}^{n} \mathcal{H}_{i} (\textbf{\^{t}}-\textbf{t}) + \alpha \mathcal{H}_{i} (\textbf{\^{r}}-\textbf{r})
\end{align}
where $\mathcal{H}$ is the Huber Loss as in (\ref{eq:hallucination_loss}). [\textbf{t}, \textbf{r}] and [\textbf{\^{t}}, \textbf{\^{r}}] are a pair of translation and rotation component for the predicted poses and the ground truth poses respectively. We use Euler angle to represent rotation since it is faster to converge as it is free from constraints unlike other representations (e.g. rotation matrix, quaternion). We also use $\alpha = 0.001$ to balance the loss between translation and rotation.

\subsection{Training Details}

The network is trained in two stages. In the first stage we train the hallucination network, while in the second stage we train the remaining networks. Note that, in the second stage, we freeze the hallucination network such that the unstable learning process in the beginning of training the other networks does not alter the learnt hallucination weights that have been trained in the first stage. We use the Adam optimizer with a 0.0001 learning rate to train the hallucination network for 200 epochs. For training the remaining networks in the second stage we employ RMSProp with a 0.001 initial learning rate, dropping by $25\%$ every 25 epochs for a total of 200 epochs. We normalize the input radiometric data by subtracting the mean over the dataset. We randomly cut the training sequence into small batches of consecutive pairs ($n = 8$) to obtain better generalization. We also sub-sample the input such that the frame rate is around 4-5 fps to provide sufficient parallax between consecutive frames. To further fine-tune the network, we alternately freeze and train the selective fusion and the pose regressor.

\section{EXPERIMENTAL RESULTS}

\subsection{Dataset}

The thermal hand-held data was collected using FLIR E95 camera at 60 fps with 464$\times$348 image resolution, while IMU data were captured using a XSens MTI-1 Series. We collected RGB-D data to train the hallucination network using Intel RealSense D435 depth camera at 30 fps and at 848$\times$480 image resolution. We place thermal and RGB-D camera with 2.5 cm distance such that it has sufficient overlap region between the thermal and RGB image. In total, we collected 19 sequences in five different buildings including a library, open office, apartment, underground storage, and an actual smoke-filled environment in firefighter training facility. We use 13 sequences for training and use the remaining data for testing. As we gathered the data mostly in public spaces, real ground truth poses are not available. Instead, we utilize VINS-Mono (visual-inertial SLAM) \cite{qin2018vins} for the comparison with the expectation that DeepTIO at least can be as accurate as visual-inertial system.

The mobile robot data was collected using a Turtlebot 2. Thermal images are captured from a Flir Boson 640 thermal camera operating at 60~fps with spatial resolution of 640$\times$512, while we utilize the same IMU device as with the hand-held data. We equip the robot with a Velodyne HDL-32E LiDAR (captures around 60,000 3D points) and an Intel RealSense Depth Camera with $680 \times 480$ RGB resolution. The distance between thermal and RGB camera is 11 cm and there is at least 2/3 spatial correspondence. In total, we have 30 sequences collected in three different buildings. We use 23 sequences for training and 7 sequences for testing. For training, we employ inertial-assisted wheel odometry (from the Turtlebot) as the pseudo ground truth. For testing, we use VICON Motion Capture system (1mm accuracy) as the ground truth for the data collected in CPS Lab while Lidar Gmapping\footnote{https://openslam-org.github.io/gmapping.html} is used for other sequences. For data collected in CPS Lab, we decorated the room with several obstacles, used different lighting condition (sufficient lighting, poor, or dark), and occasionally had a person walking in the camera frame.

\subsection{Evaluation Metrics}
To evaluate the proposed model, we utilize Mean Square (MS) of Relative Pose Error (RPE) and Absolute Trajectory Error (ATE), since they have been widely used for measuring VO or visual SLAM accuracy \cite{sturm2012benchmark}. As we have different (pseudo) ground truth format to compare with (VICON, Lidar Gmapping, or VINS-Mono), we align the predicted poses with the (pseudo) ground truth using Horn approaches and evaluate only the poses that are closest in time. We use the evaluation tools from TUM RGB-D dataset to do this\footnote{https://vision.in.tum.de/data/datasets/rgbd-dataset}.

\subsection{Sensitivity Analysis}

To understand the influence of the hallucination network, we perform a sensitivity analysis in the following section.
\begin{figure}
    \centering
    \includegraphics[width=0.8\columnwidth]{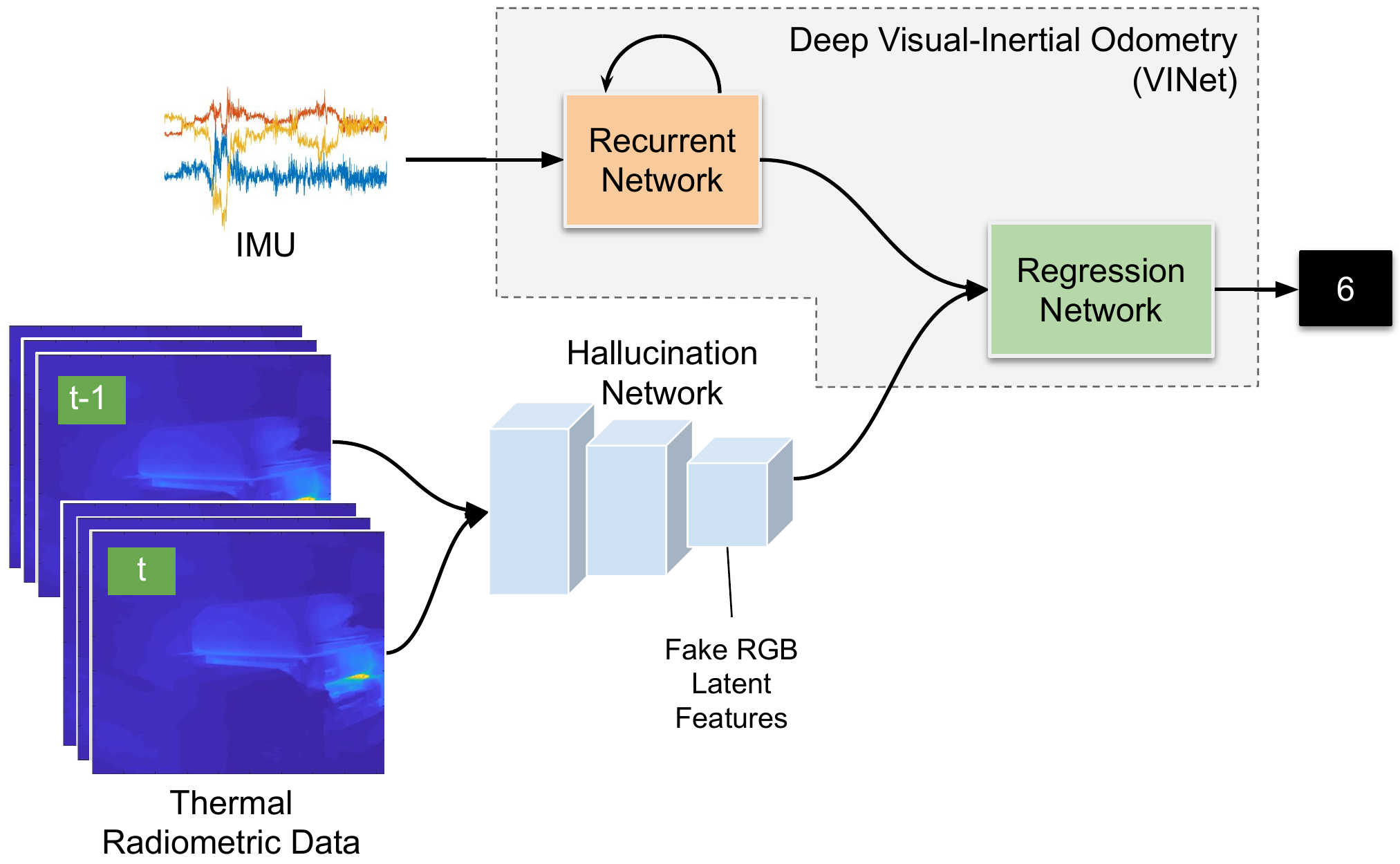}
    \caption{The hallucination network is validated by feeding the fake RGB features to VINet and measuring the pose estimation discrepancy.}
\label{fig:deeptio_validation}
\end{figure}

\begin{figure}[!ht]
    \centering
    \includegraphics[width=4.2cm,trim=0.2cm .1cm .2cm .7cm,clip]{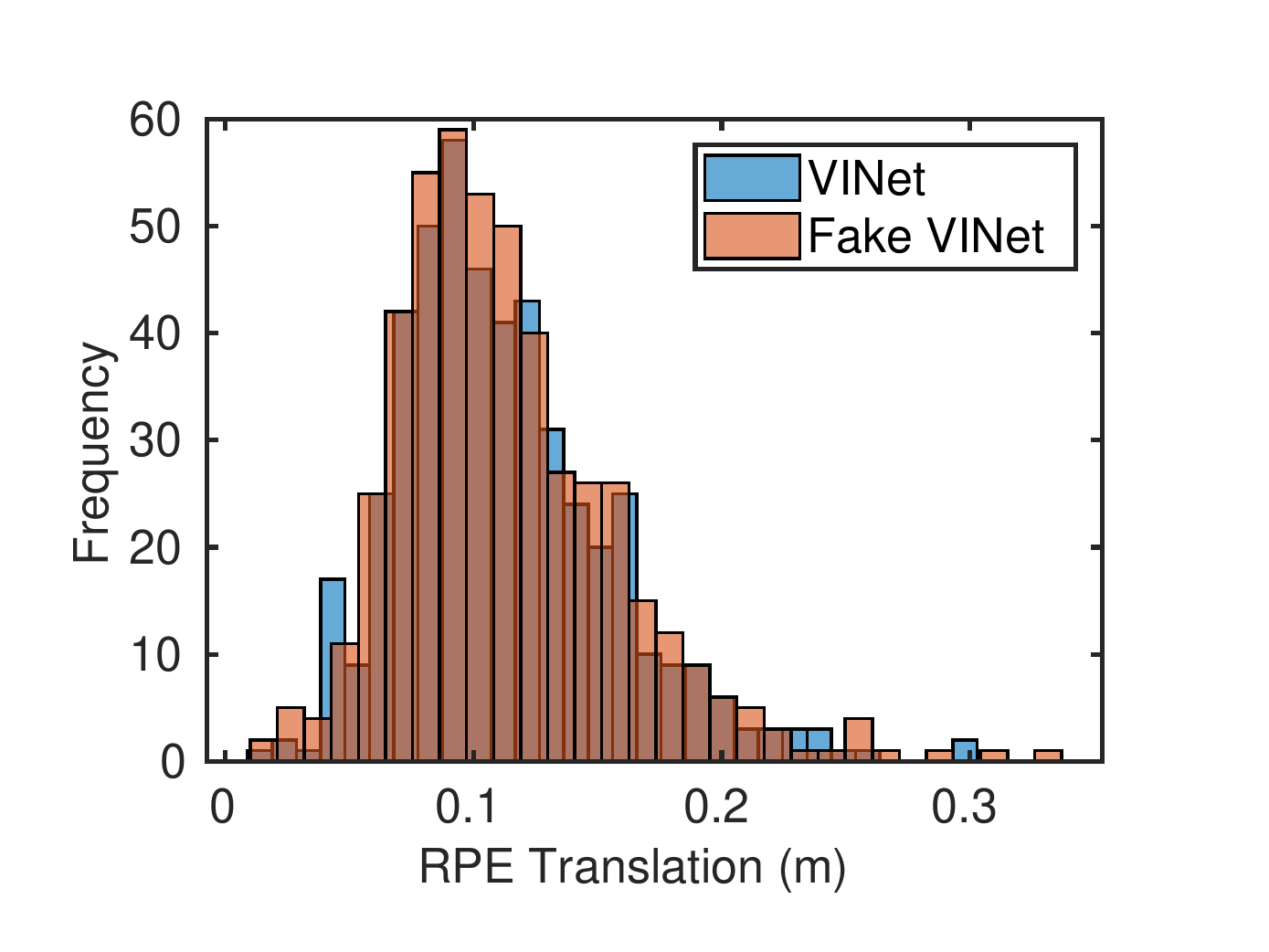}
    \includegraphics[width=4.2cm,trim=0.2cm .1cm .2cm .7cm,clip]{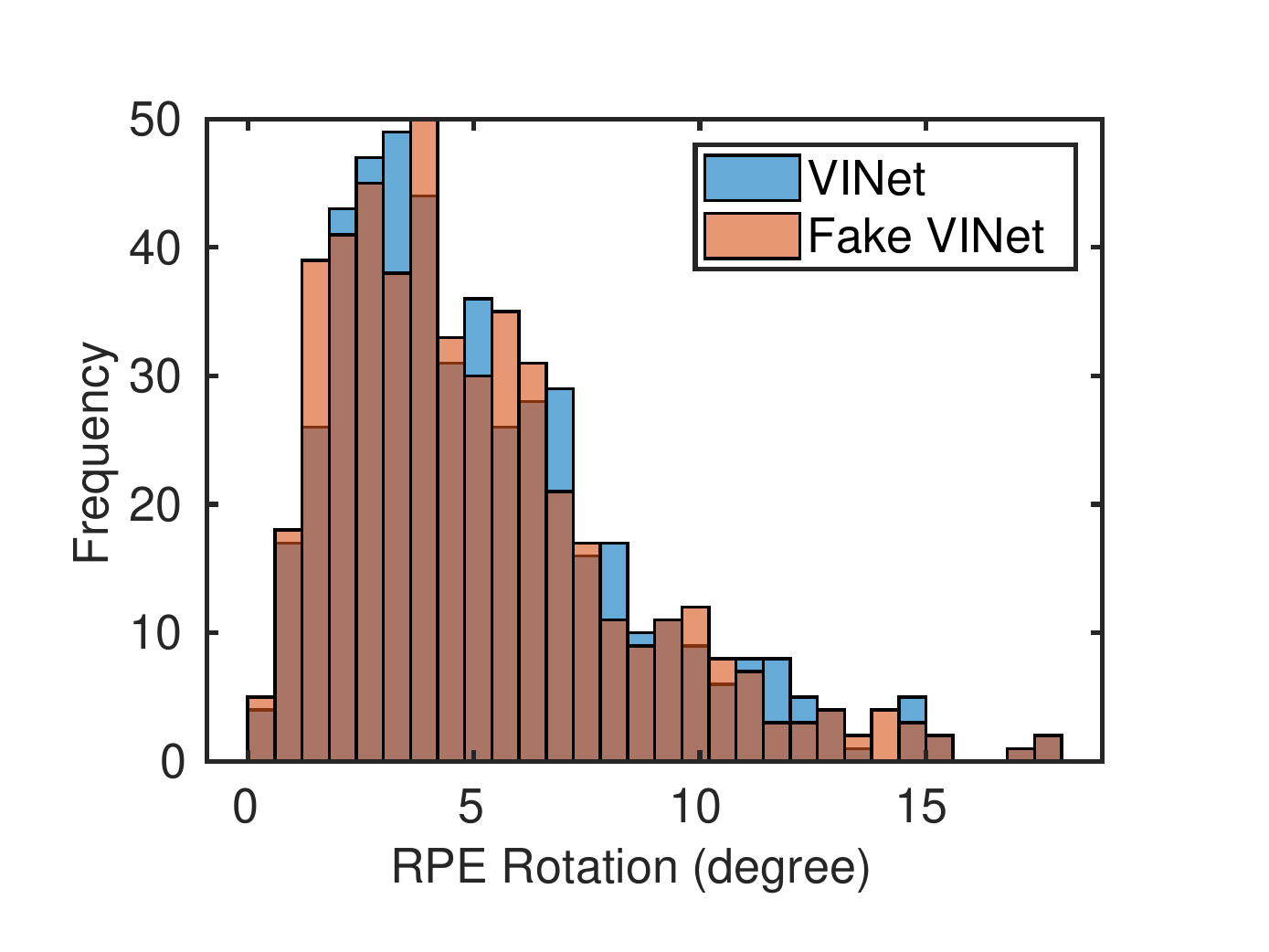}
    \caption{Relative Pose Error (RPE) distribution between VINet and Fake VINet for both translation and rotation.}
\label{fig:rpe_validate_hallu}
\end{figure}

\subsubsection{Validating the Hallucination Network}
\begin{figure*}[!ht]
    \centering
    \includegraphics[width=6.7cm,trim=0.9cm 3.4cm 0.6cm 2.5cm,clip]{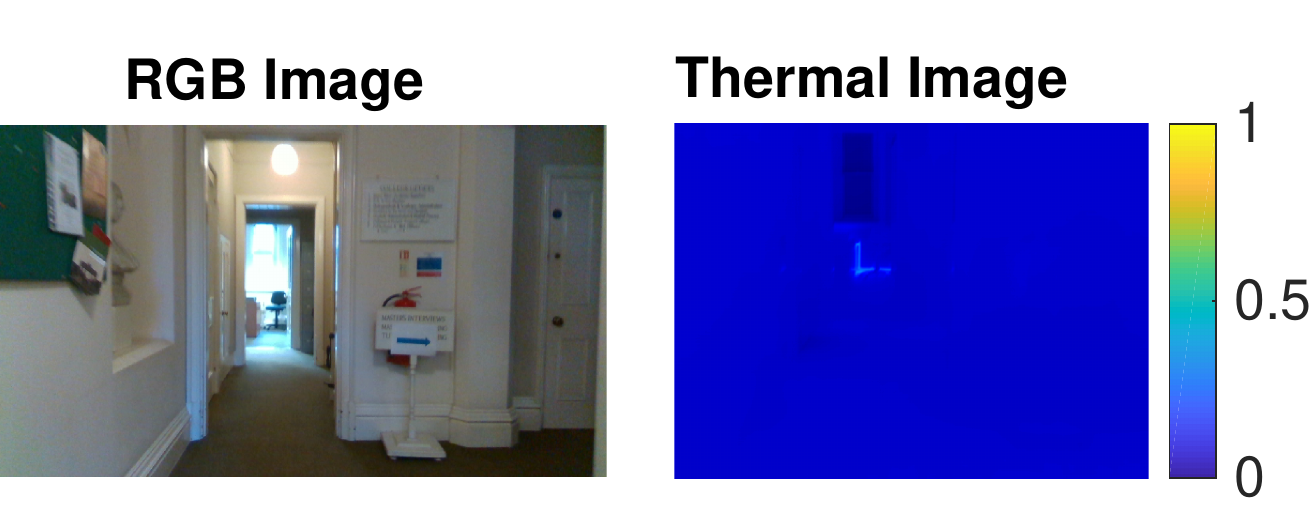}
    \includegraphics[width=9cm,trim=0.7cm 4.cm 1.5cm 3.5cm,clip]{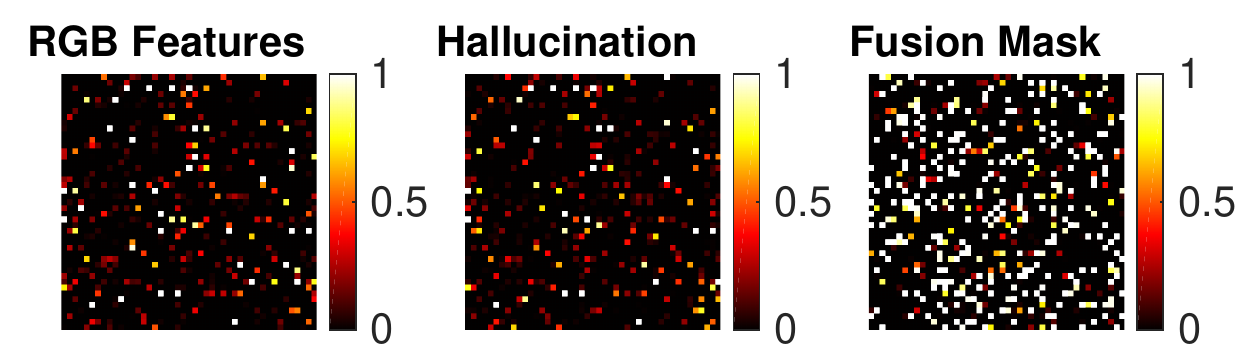} \\
    \includegraphics[width=6.7cm,trim=0.9cm 3.4cm 0.6cm 3.2cm,clip]{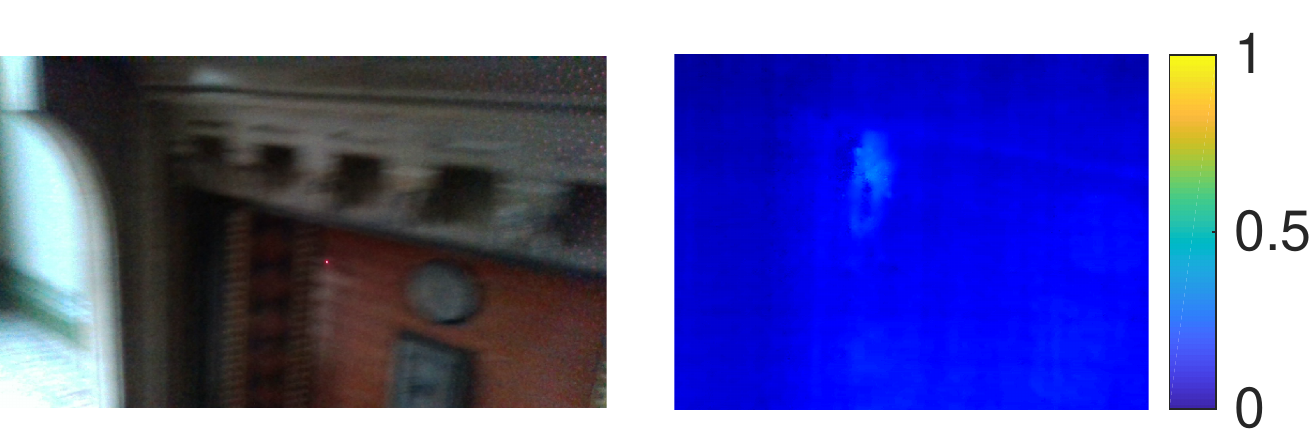}
    \includegraphics[width=9cm,trim=0.7cm 4.cm 1.5cm 3.9cm,clip]{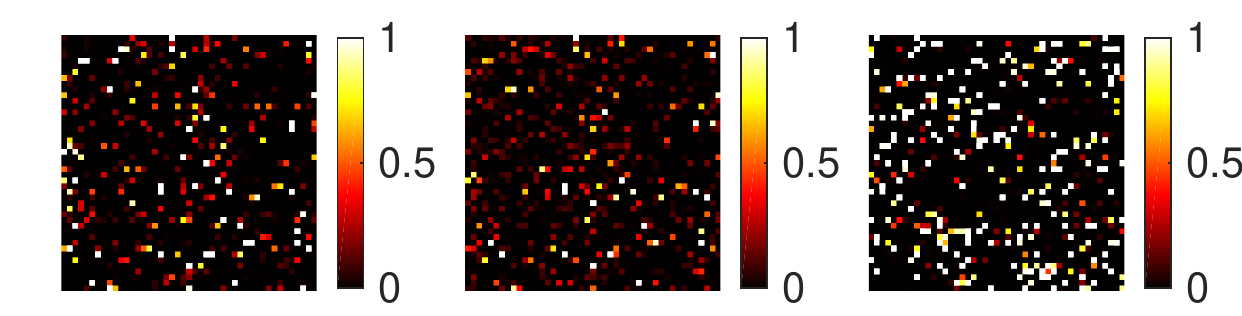}
    \caption{Comparison between RGB features produced by VINet and fake RGB features produced by DeepTIO's hallucination network in Corridor 2. Top: example of accurate hallucination. Bottom: example of erroneous hallucination. From left to right: RGB image, thermal image, original RGB features, hallucinated RGB features, and fusion mask for the hallucination features generated by DeepTIO.}
\label{fig:features_visualization}
\end{figure*}

To validate the hallucination network, we replace the visual decoder network from VINet with the hallucination network from DeepTIO as seen in Fig. \ref{fig:deeptio_validation}. By feeding the hallucinated visual features (fake RGB features) to the original VINet, we can measure how accurate the learnt representation produced by the hallucination network are. Fig. \ref{fig:rpe_validate_hallu} shows the distribution of RPE between VINet and Fake VINet (VINet with input from fake RGB features). It can be seen that the error distribution for both translation and rotation are very similar, showing the success of training the hallucination network. Table \ref{table:validate_hallu} shows how close the average RPE between VINet and Fake VINet are. Surprisingly, the Fake VINet got a slightly better result for rotation estimation, showing the efficacy of training using Huber Loss. Fig. \ref{fig:features_visualization} illustrates the visualization of the output features from VINet and from hallucination network in the test sequence. It can be seen that the network can hallucinate visual features accurately (top) despite the lack of features in thermal domain. We presume that as long as there are spatial correspondences between thermal and RGB image, the network will learn to associate similar features and interpolate the missing ones. However, there are also cases when the hallucination network produces erroneous features (bottom) due to blurriness or lack of thermal edges. In this case, selective fusion plays important roles for selecting only relevant information from the hallucination network. It can be seen that in the erroneous case example, the DeepTIO's selective fusion produces less dense fusion masks, indicating less features are being used.

\begin{table}[t]
\centering
\begin{threeparttable}
  \caption{RPE between VINet and Fake VINet}
  \renewcommand{\arraystretch}{0.9}
  \fontsize{9}{12}\selectfont
    \label{table:validate_hallu}
  \begin{tabular}{lcc}
    \hline
    \noalign{\smallskip}
    Model & $\textbf{t}$ (m) & $\textbf{r}$ ($^{\circ}$) \\
    \hline
    \noalign{\smallskip}
    VINet & 0.1124 & 5.1954 \\
    \hline
    Fake VINet ($\mathcal{L}_2$) & 0.1197 & 5.1926 \\
    Fake VINet (Huber) & \textbf{0.1128} & \textbf{5.0739} \\
    \hline
  \end{tabular}
\end{threeparttable}
\end{table}

\subsubsection{The Influence of Each Feature Modality}
To understand the influence of each feature modality, we decouple each feature modality, train it separately, and test the result. The result can be seen in Table \ref{table:influence_feature} and shows that thermal alone got the worst accuracy, implying the difficulty of estimating odometry solely based on temperature profile information. IMU alone clearly shows much stronger performance although the optimal solution may require to produce only 3-DoF poses (instead of 6-DoF) as seen in \cite{chen2018ionet} since there is not enough information from the IMU data to produce accurate 6-DoF poses. Incorporating IMU with thermal features or fake RGB features improves the accuracy (ATE) as the thermal or the visual features constraints the IMU error growth. Adding Fake RGB features to the model with IMU+Thermal further reduces the ATE, indicating that the hallucinated visual features help generate more accurate poses. Note that all feature fusions (with the same mark in Table \ref{table:influence_feature}) have the same network capacity, indicating that the improved accuracy is due to more useful information, rather than increased network capacity.

\subsubsection{The Influence of Selective Fusion}
As seen in Table \ref{table:influence_feature}, incorporating selective fusion to the combined features consistently reduces the ATE over the network without selective fusion. This shows that selective fusion plays important role in producing accurate results as each feature modality comes with intrinsic noises, the hallucination network may produce erroneous visual features, and there is time misalignment between the sensors and the ground truth. Finally, putting together all feature modalities with selective fusion yields the strongest performance for both RPE and ATE.

\begin{table}[t]
\centering
\begin{threeparttable}
  \caption{The Impact of Each Feature Modality and Selective Fusion}
  \renewcommand{\arraystretch}{0.9}
  \fontsize{9}{12}\selectfont 
    \label{table:influence_feature}
  \setlength\tabcolsep{4pt} 
  \renewcommand{\arraystretch}{0.95}
  \begin{tabular}{lccc|c}
    \hline
    Features & SF$^\dag$ & $\textbf{t}$ (m) & $\textbf{r}$ ($^{\circ}$) & ATE (m) \\
    \hline
    Thermal & - & 0.1497  & 6.5839 & 6.8347 \\
    IMU & - & 0.1204 & 5.0151 & 1.7779 \\
    IMU+Thermal$^*$ & - & 0.1133 & 5.3112 & 1.4731 \\
    IMU+Thermal$^+$ & \checkmark & 0.1192 & 5.0461 & 0.7122 \\
    IMU+Fake RGB$^*$ & - & 0.1153 & 5.2378 & 1.5021  \\
    IMU+Fake RGB$^+$ & \checkmark & 0.1090 & 5.2300 & 1.0280 \\
    IMU+Thermal+Fake RGB$^*$ & - & 0.1080 & 5.2127 & 1.2824 \\
    IMU+Thermal+Fake RGB$^+$ & \checkmark & \textbf{0.1074} & \textbf{4.8826} & \textbf{0.5267} \\
    \hline
    \multicolumn{5}{p{220pt}}{\footnotesize $^*$ has 52 M weights, while $^+$ has 136 M weights.}\\
    \multicolumn{5}{p{200pt}}{\footnotesize \dag  Whether Selective Fusion (SF) is employed or not.} \end{tabular}
\end{threeparttable}
\end{table}

\begin{figure*}[!ht]
    \centering
    \subfloat[Hand-held: Corridor 1]{
        \includegraphics[width=4.5cm,trim=1cm .5cm 1cm .5cm,clip]{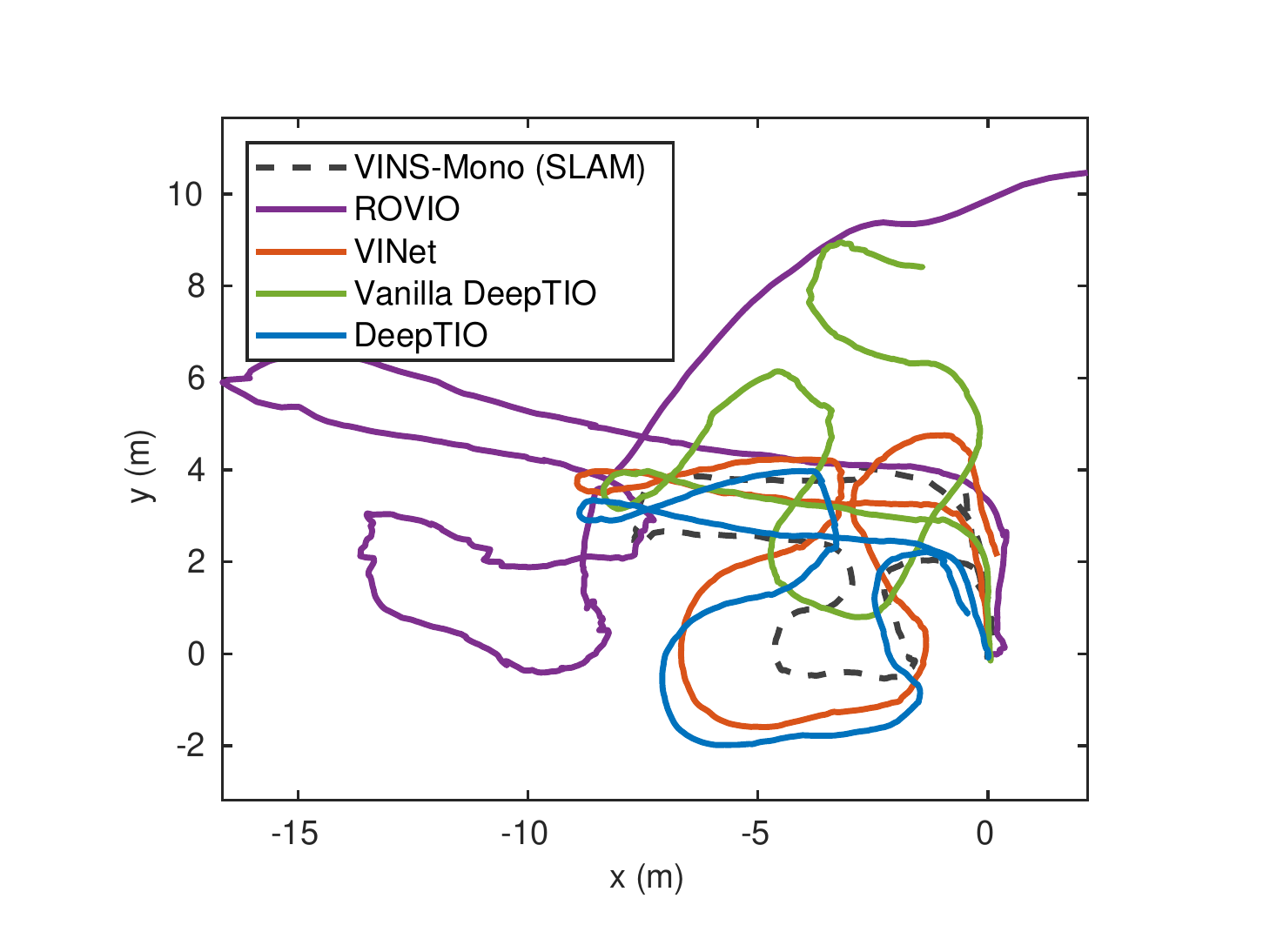}
        \hspace{-0.5cm}}
    \subfloat[Hand-held: Corridor 2]{
        \includegraphics[width=4.5cm,trim=1cm .5cm 1cm .5cm,clip]{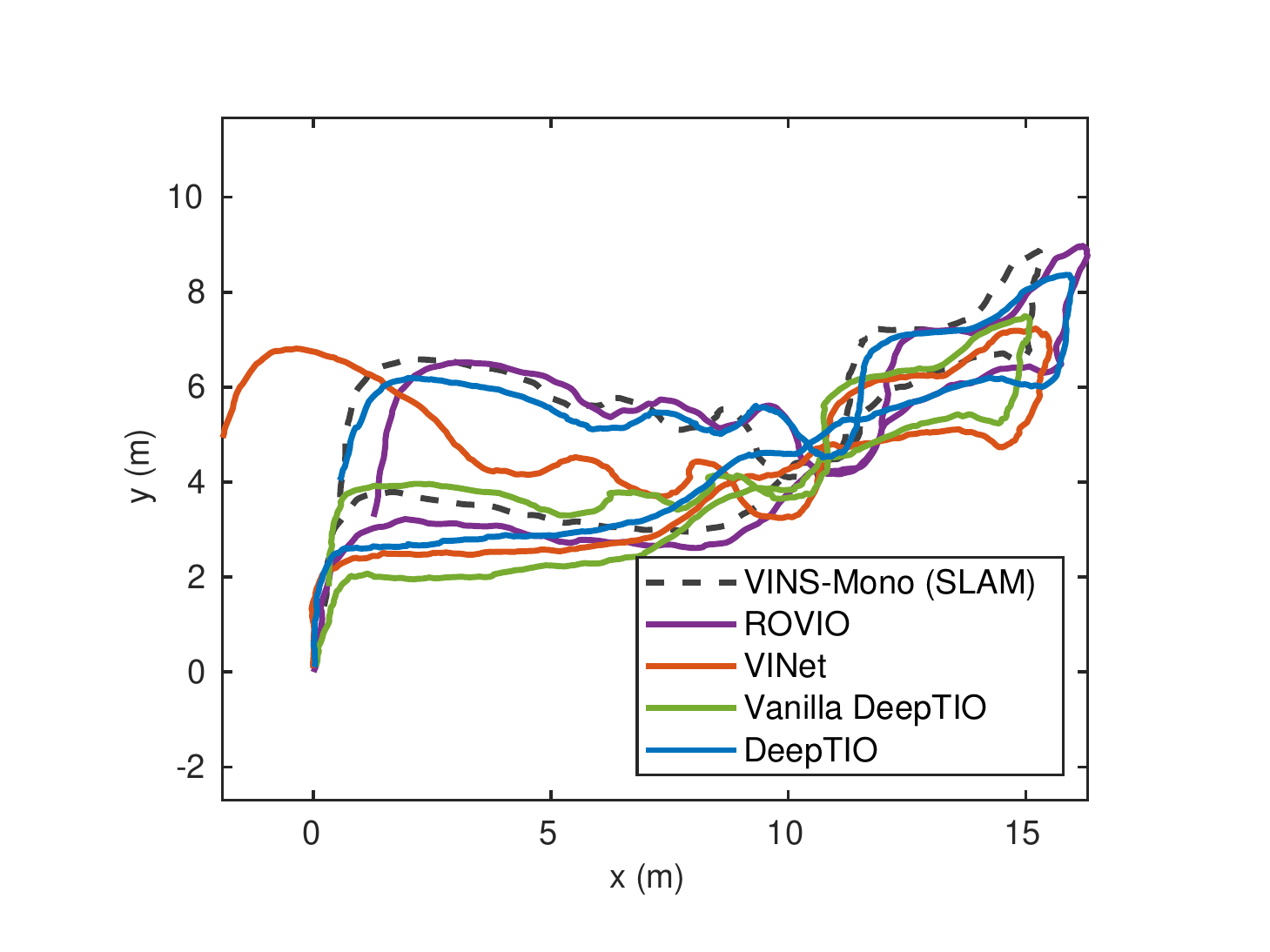}
        \hspace{-0.5cm}}
    \subfloat[Hand-held: Library 2]{
        \includegraphics[width=4.5cm,trim=1cm .5cm 1cm .5cm,clip]{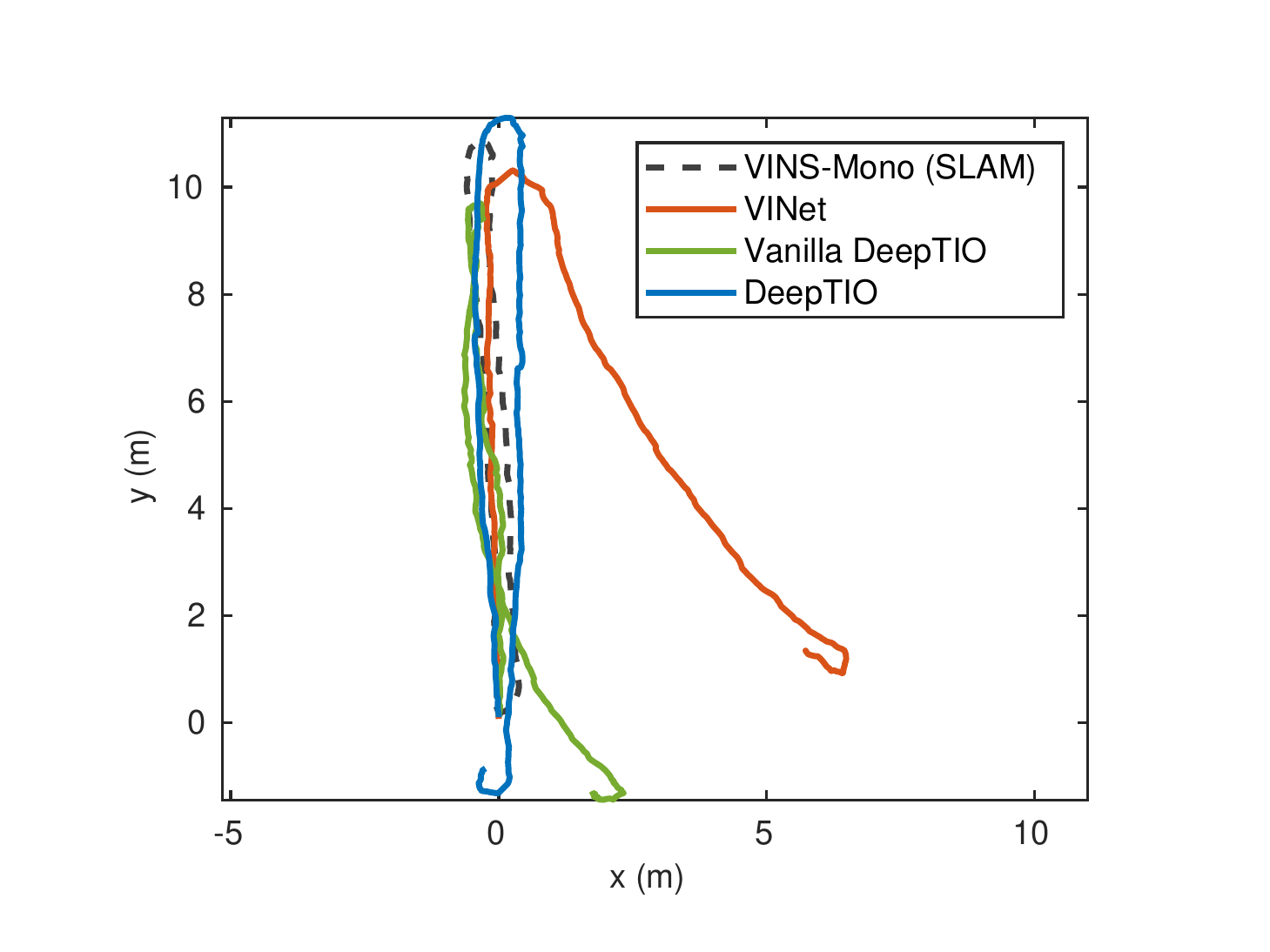}
        \hspace{-0.5cm}}
    \subfloat[Hand-held: Firefighter facility]{
        \vspace{-0.9cm}
        \includegraphics[width=5cm,trim=1cm .5cm 1cm 5.5cm,clip]{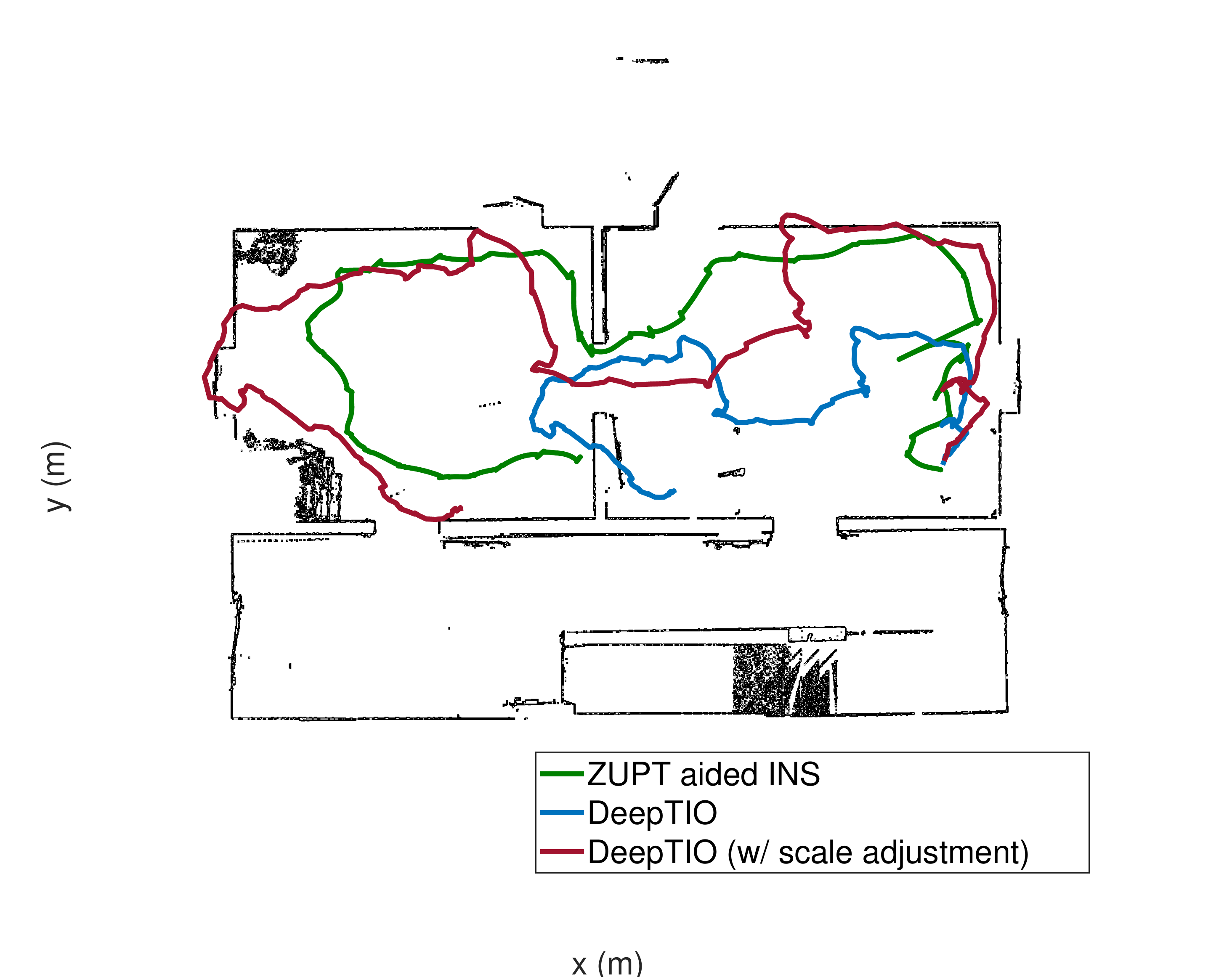}
        \hspace{-0.5cm}}
    \\ \vspace{-0.5cm}
    \subfloat[Turtlebot: CPS Lab 1]{
        \vspace{-0.9cm}
        \hspace{-0.5cm}
        \includegraphics[width=4.5cm,trim=1cm .5cm 1cm .3cm,clip]{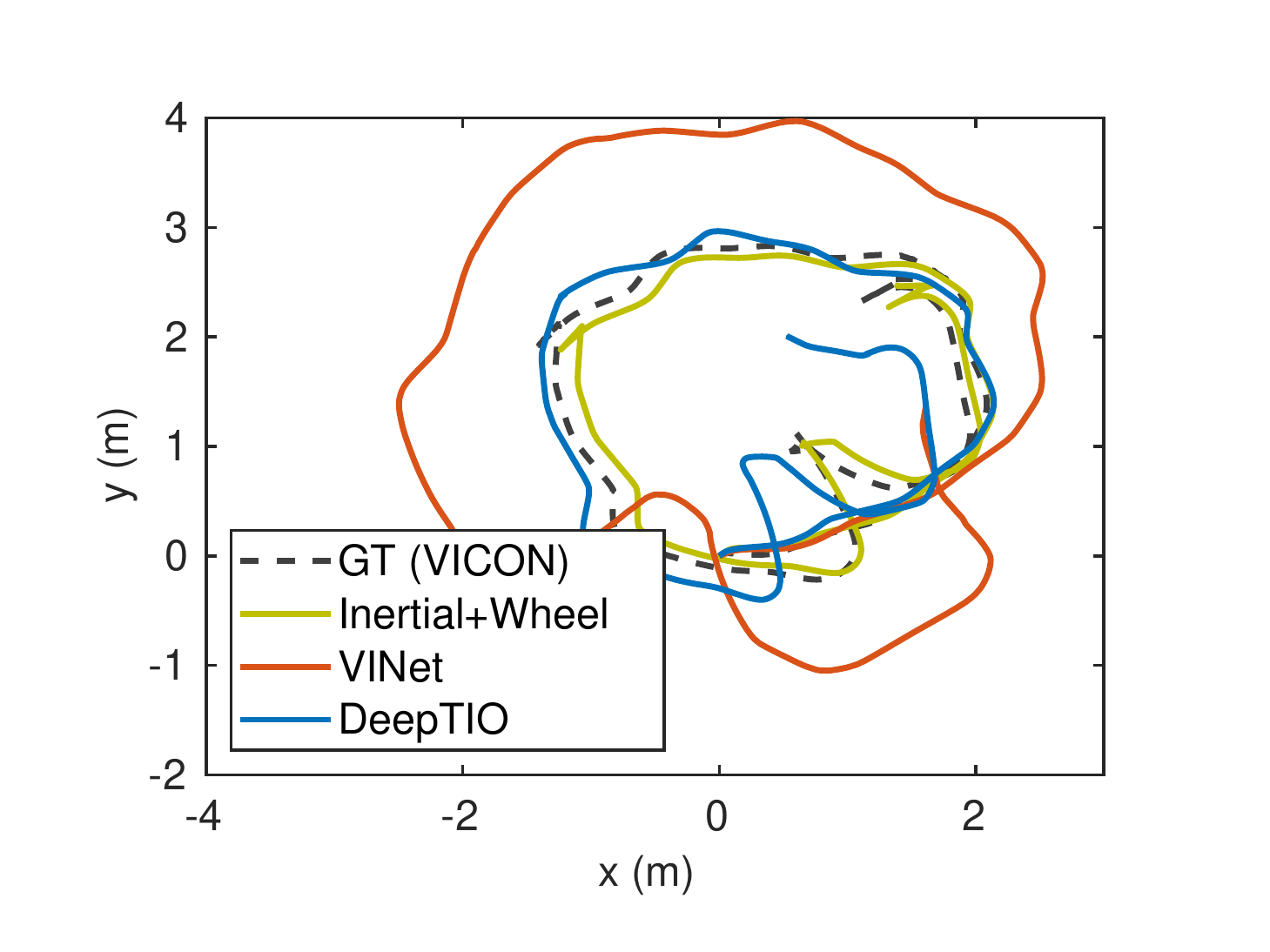}
        \hspace{-0.5cm}}
    \subfloat[Turtlebot: CPS Lab 4 (dark)]{
        \vspace{-0.9cm}
        \includegraphics[width=4.5cm,trim=1cm .5cm 1cm .3cm,clip]{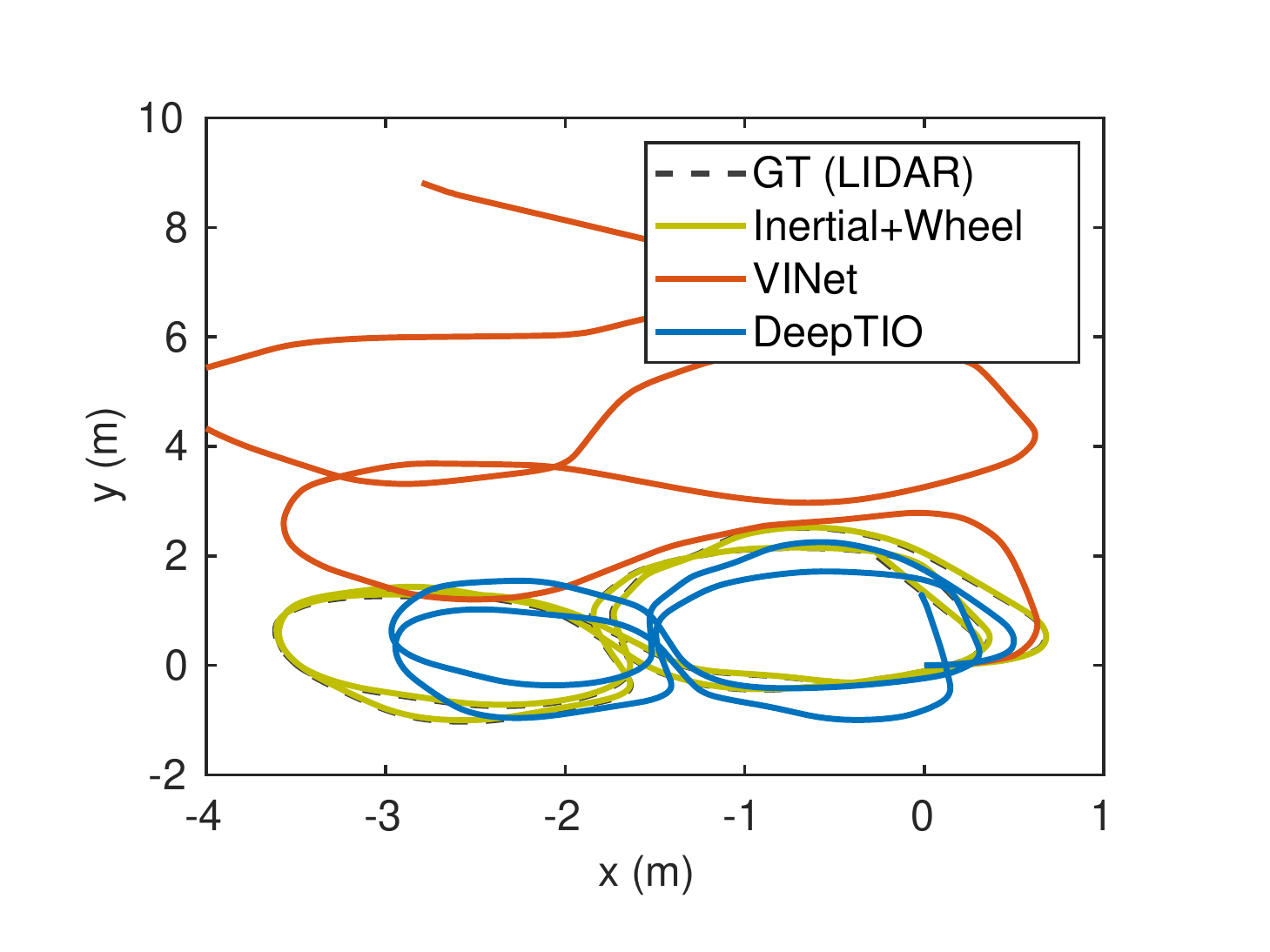}
        \hspace{-0.5cm}}
    \subfloat[Turtlebot: Corridor 1]{
        \vspace{-0.9cm}
        \includegraphics[width=4.5cm,trim=1cm .5cm 1cm .3cm,clip]{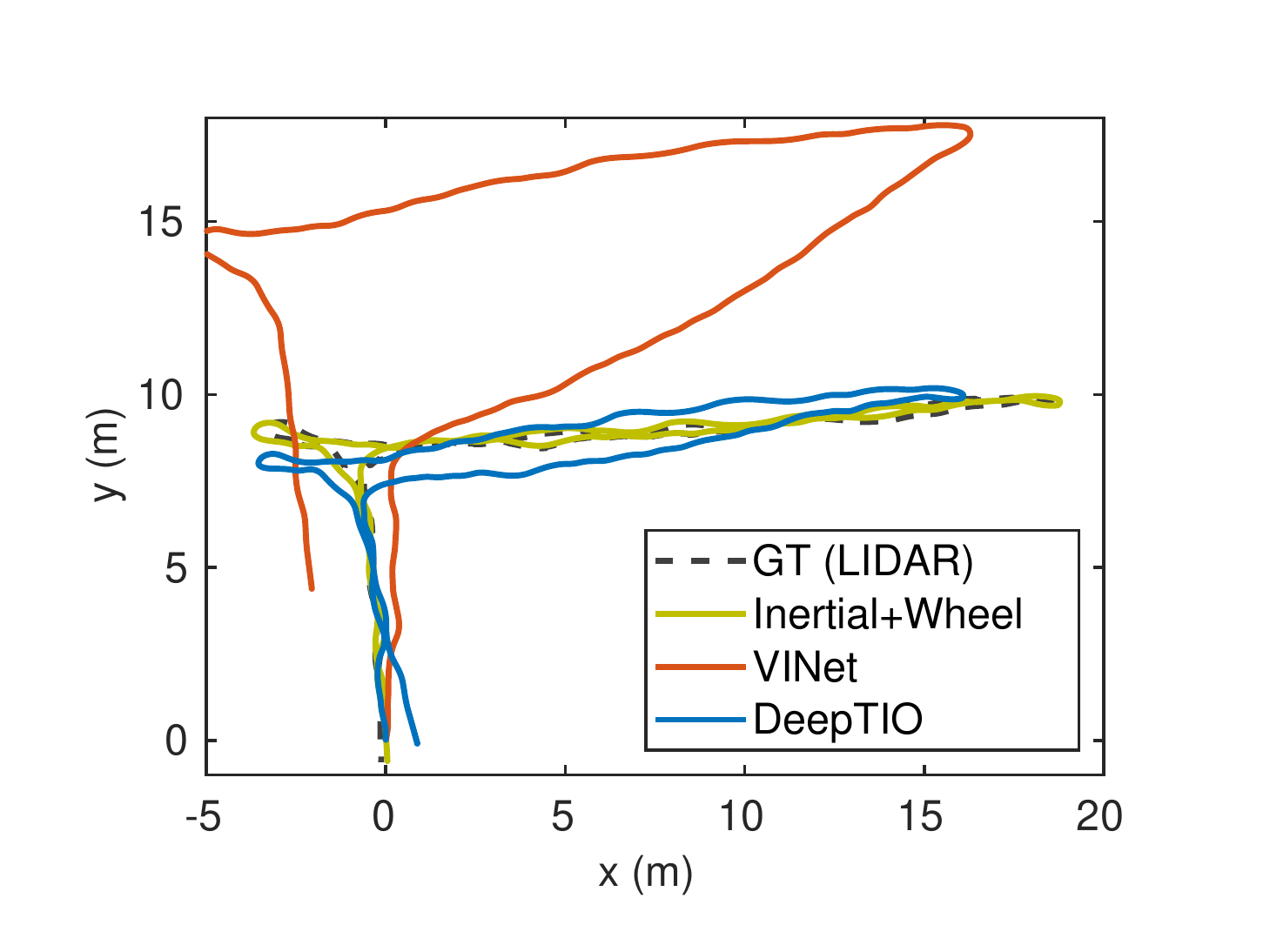}
        \hspace{-0.5cm}}
    \subfloat[Turtlebot: Corridor 2]{
        \vspace{-0.9cm}
        \includegraphics[width=4.5cm,trim=1cm .5cm 1cm .3cm,clip]{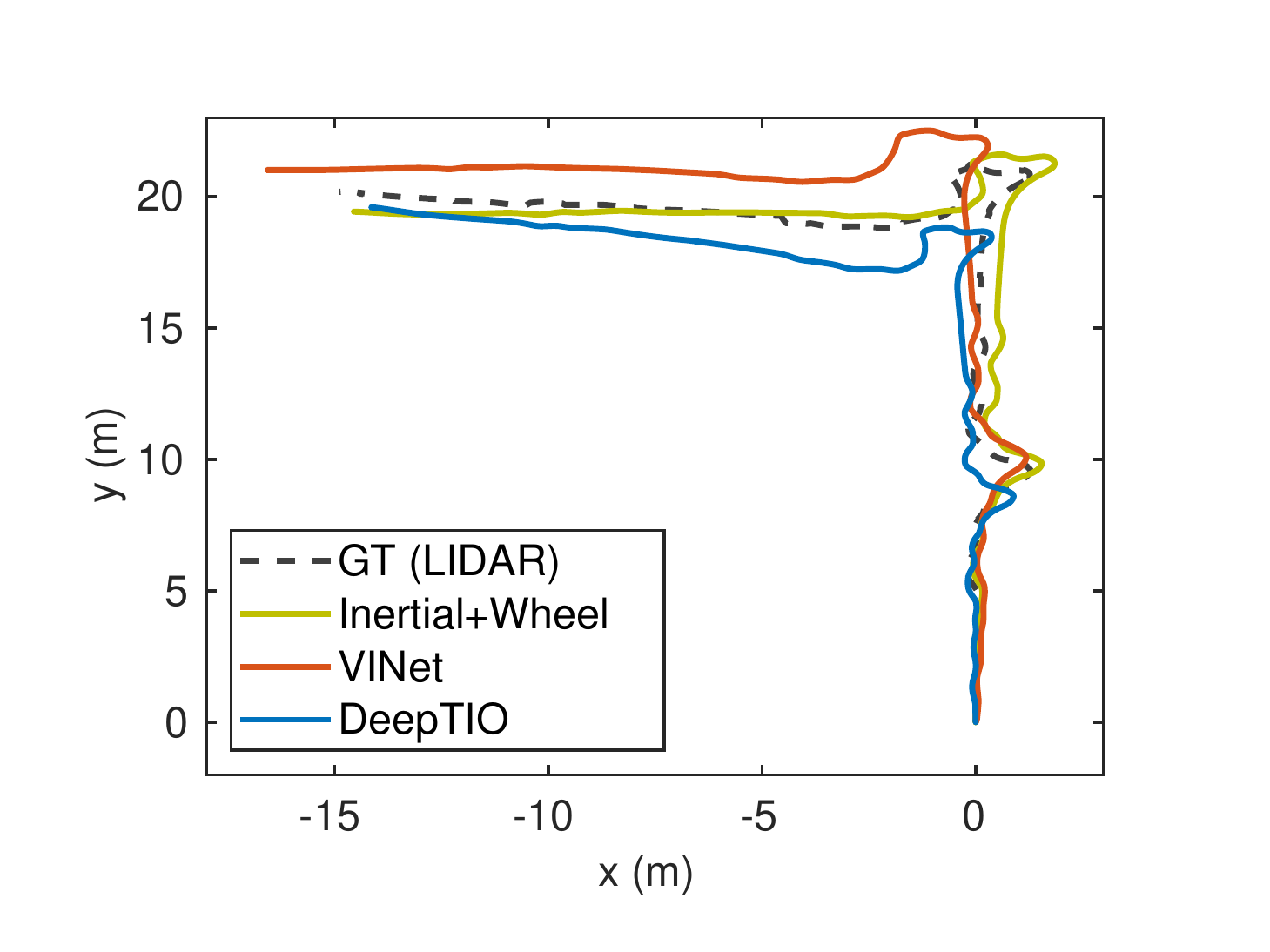}
        \hspace{-0.5cm}}
    \caption{Qualitative evaluation in test sequences. (a)-(d) are test with hand-held data. (e)-(h) are test with mobile robot data. (d) is test in real emergency scenario with smoke-filled environment. We qualitatively compared DeepTIO with ZUPT aided INS as VIO, VI-SLAM, or even Lidar odometry does not work in this visually-denied scenario.}
\label{fig:trajectories}
\end{figure*}

\subsection{Evaluation on Hand-held Data}

\subsubsection{Test in Benign Environment} 
We test our model across different buildings and compare it with the state-of-the-art VIO frameworks to show that our DeepTIO solution is comparable. For conventional approach we employ ROVIO \cite{bloesch2015robust}, which tightly fuses IMU and visual data with an iterated extended Kalman filter. For deep learning based approaches, we use VINet \cite{clark2017vinet} which fuses IMU and visual features in the intermediate layer. Both ROVIO and VINet uses RGB as the input since it easily lose tracks when we use thermal as the input. We also compare with Vanilla DeepTIO, a version of DeepTIO without the visual hallucination network. Fig. \ref{fig:trajectories} (a)-(d) depicts the qualitative results in this scenario.

Table \ref{table:evaluation_hand_held} shows the numerical evaluation results in terms of ATE. ROVIO provides good accuracy in Corridor 2, although it suffers from large scaling problem and loses tracking in Corridor 1 due to lack of visual features when the camera faces white, flat walls. 
In misaligned sequences, ROVIO completely fails to initialize, since it requires tightly synchronized inputs. 
VINet also performs well when good alignment is available but suffers from large drift in presence of time misalignment. This shows that directly concatenating features may lead to sub-optimal performances. Nevertheless, VINet can still produce odometry where ROVIO completely fails, showing that deep learning approaches are more robust against sensor alignment issue. However, the best results are achieved by Vanilla DeepTIO and DeepTIO as they employ selective fusion which is proven to be robust to time synchronization issues \cite{chen2019selective}. Note that Vanilla DeepTIO and DeepTIO use a smaller thermal image resolution (464x348) compared to the RGB images used by ROVIO and VINet (848x480).

Vanilla DeepTIO achieves excellent results in Corridor 2, Large Office, and Library 2, but suffers from drift in Corridor 1 and Library 1. DeepTIO, on the other hand, produces better results due to the additional information provided by the hallucination network. Nonetheless, estimating an accurate scale is a problem in some sequences. As seen in the Large Office sequence, both Vanilla DeepTIO and DeepTIO give inaccurate scale, possibly due to a large variation of walking speeds. This scaling problem is very common in VO or VIO (as seen in ROVIO test in Corridor 1) and remains an open problem. Overall, DeepTIO yields the best ATE against the competing approaches, with an average ATE of $1.67$ m.

\begin{table}[t]
\centering
\begin{threeparttable}
    \setlength{\tabcolsep}{5pt} 
  \caption{ATE (m) For Experiment with Hand-held Data}
  \renewcommand{\arraystretch}{0.9}
  \setlength\tabcolsep{4pt}
  \fontsize{9}{12}\selectfont
  \label{table:evaluation_hand_held}
  \begin{tabular}{lcccc}
    \hline
    \noalign{\smallskip}
    & \multirow{2}*{ROVIO} &\multirow{2}*{VINet} & Vanilla & DeepTIO \\
    & & & DeepTIO (\textbf{ours}) &  (\textbf{ours}) \\
    \hline
    \noalign{\smallskip}
    Corridor 1 & 6.2496 & \textbf{1.8825} & 2.1975 & 1.9333 \\
    Corridor 2 & \textbf{0.3343} & 1.1036 & 0.7122 & 0.5267 \\
    Large Office* & failed & 4.4359 & 3.3088 & \textbf{3.2648} \\
    Library 1* & failed & 5.2647 & 2.5698 & \textbf{2.0532} \\
    Library 2* & failed & 1.6812 & 1.5741 & \textbf{0.5735} \\
    \hline
    Mean & 3.2920 & 2.8736 & 1.7502 & \textbf{1.6703} \\
    \hline
    \multicolumn{5}{p{210pt}}{\footnotesize *There is time misalignment among sensors for about 1-2 second.}
  \end{tabular}
\end{threeparttable}
\end{table}

\subsubsection{Test in Smoke-filled Environment}

In the smoke-filled environment, none of the VIO frameworks can work as the RGB camera only captures black frames. Even Lidar odometry does not work well as near-visible light is blocked by the smoke \cite{bijelic2019seeing}. In this case, we cannot provide quantitative evaluation with any (pseudo) ground truth. We instead provide a qualitative comparison with a zero-velocity-aided Inertial Navigation System (INS) \cite{wahlstrom2019zero}, which is not impacted by visibility. This navigation system utilizes foot-mounted inertial sensors to detect Zero Velocity Updates (ZUPT) and thereby mitigates the fast error growth of stand-alone inertial navigation. Fig. \ref{fig:trajectories} (d) shows the output trajectory together with the floor plan generated by FARO Lidar collected before the experiment with smoke. It can be seen that DeepTIO yields a similar trajectory shape to ZUPT. This shows that our model, despite being trained in a benign environment, can generalize to a smoke-filled environment as the thermal camera is not affected by the smoke. However, there is scaling issue which probably due to different speed of the camera (as an effect of different walking speed) or different temperature profile compared to the one observed in the training data. If we adjust the scale of DeepTIO, it can be seen that the prediction is very close to ZUPT. This shows that our model is promising for odometry estimation in smoke-filled environments.

\subsubsection{Memory and Execution Time}
The network was trained on an NVIDIA TITAN V GPU and required around 6-18 hours for training the hallucination network and 6-20 hours to train the remaining networks. The network contains around 136 millions weights, requiring 847 MB of space. Neglecting the time to load and normalize the input, the model can run at 40 fps on a TITAN V and 5 fps on a standard CPU.

\begin{figure}[!ht]
    \centering
    \includegraphics[width=7cm,trim=0.9cm 0.5cm .5cm 1.cm,clip]{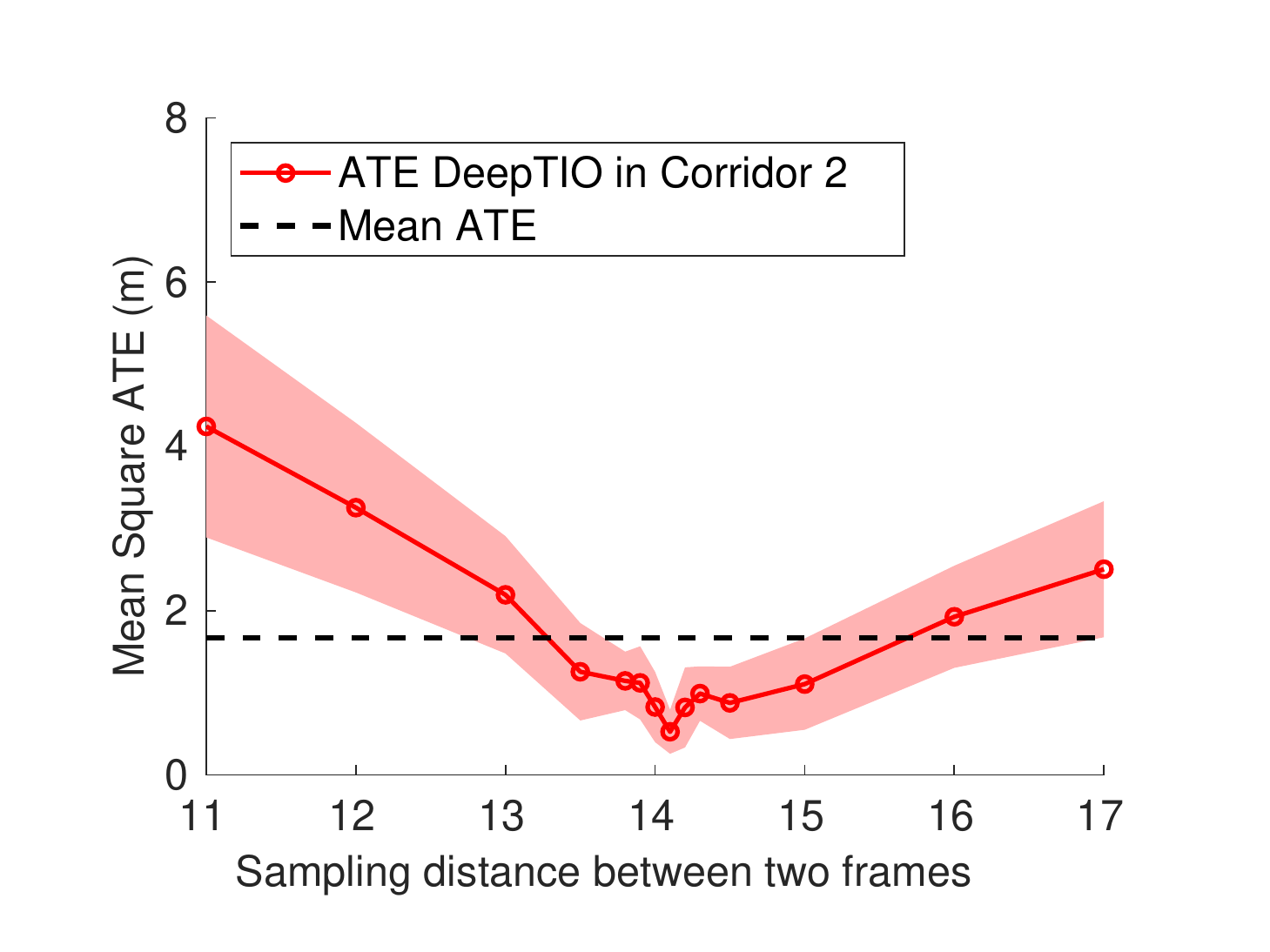}
    \caption{Sensitivity towards sampling rate (fps). By using 14 sampling distance between two frames (optimum performance), it keeps the prediction rate around 4.2 (for 60 thermal fps), which is still in the range of 4-5 fps used during training.}
\label{fig:sensitivity_fps}
\end{figure}

\subsection{Evaluation on Mobile Robot Data}
Fig. \ref{fig:trajectories} (e)-(h) depicts the test results of mobile robot data collected in two different buildings. As can be seen, DeepTIO yields very accurate result in all test sequences, outperforming VINet and generating a similar trajectory to inertial-assisted wheel odometry which we use as pseudo ground truth in training. In the sequence with no lighting (Fig. \ref{fig:trajectories} (f)), VINet experiences significant drift due to the low RGB illumination. On the contrary, DeepTIO can produce an accurate trajectory despite varying lighting conditions. DeepTIO also works very well in long sequences with challenging motions involving U-turns and with people walking within the camera frame as can be seen in Fig. \ref{fig:trajectories} (g). It indicates that using three modalities (although the RGB is a fake one) can yield more robust predictions in challenged environments compared to just using two modalities (e.g. VINet).

Table \ref{table:evaluation_robot} describes the quantitative result for all test sequences. DeepTIO achieves a very low ATE (around 0.5 m in average). Comparing this result with the hand-held data, it implies that learning mobile robot odometry is easier than learning hand-held odometry. This is most likely because the robot movement is more stable in terms of speed and planar constraints.

\begin{table}[t]
\centering
\begin{threeparttable}
    \setlength{\tabcolsep}{5pt} 
  \caption{ATE (m) For Experiment with Mobile Robot Data}
  \renewcommand{\arraystretch}{0.9}
  \setlength\tabcolsep{6pt}
  \fontsize{9}{12}\selectfont
  \label{table:evaluation_robot}
  \begin{tabular}{lcc|c}
    \hline
    \noalign{\smallskip}
    &\multirow{2}*{VINet} & DeepTIO & Inertial \\
    & & (\textbf{ours}) & +Wheel \\
    \hline
    \noalign{\smallskip}
    CPS Lab 1 & 1.3619 & \textbf{0.3824} & 0.0931 \\
    CPS Lab 2 & 1.1440 & \textbf{0.2785} & 0.0805 \\
    CPS Lab 3 (poor) & 1.2225 & \textbf{0.2250} & 0.0033 \\
    CPS Lab 4 (dark) & 1.8379 & \textbf{0.3807} & 0.0036 \\
    Corridor 1 & 2.4497 & \textbf{0.8988} & 0.1279 \\
    Corridor 2 & 0.8294 & \textbf{0.7433} & 0.3595 \\
    Corridor 3 & 1.4629 & \textbf{0.7307} & 0.2297 \\
    \hline
    Mean & 1.4726 & \textbf{0.5199} & 0.1282 \\
    \hline
  \end{tabular}
\end{threeparttable}
\end{table}

\subsection{Limitations}
Despite the fact that DeepTIO can work well in our test scenarios, it is very sensitive to the image sampling rate. As we trained DeepTIO with a frame rate of 4-5 fps, the network will only perform well when using this frame rate. When inferring with lower or faster fps, the accuracy will degrade as seen in Fig. \ref{fig:sensitivity_fps}. Training with multiple fps at the same time might be possible to obtain robustness against different sampling rates. This may also alleviate the problem of scaling as the network would be trained with more variations of parallax.


\section{CONCLUSIONS}

We presented a novel DNN-based method for thermal-inertial odometry (DeepTIO) using hallucination networks. We demonstrated that the hallucination network can provide side information for the thermal network and, combined with a selective fusion mechanism, is able to produce accurate odometry estimation. Future works include the development of loop closure detection or a full thermal SLAM system to alleviate the drift problem in longer sequences.

\ifCLASSOPTIONcaptionsoff
  \newpage
\fi



%


\bibliography{library}

\begin{thebibliography}{10}
\providecommand{\url}[1]{#1}
\csname url@samestyle\endcsname
\providecommand{\newblock}{\relax}
\providecommand{\bibinfo}[2]{#2}
\providecommand{\BIBentrySTDinterwordspacing}{\spaceskip=0pt\relax}
\providecommand{\BIBentryALTinterwordstretchfactor}{4}
\providecommand{\BIBentryALTinterwordspacing}{\spaceskip=\fontdimen2\font plus
\BIBentryALTinterwordstretchfactor\fontdimen3\font minus
  \fontdimen4\font\relax}
\providecommand{\BIBforeignlanguage}[2]{{%
\expandafter\ifx\csname l@#1\endcsname\relax
\typeout{** WARNING: IEEEtran.bst: No hyphenation pattern has been}%
\typeout{** loaded for the language `#1'. Using the pattern for}%
\typeout{** the default language instead.}%
\else
\language=\csname l@#1\endcsname
\fi
#2}}
\providecommand{\BIBdecl}{\relax}
\BIBdecl

\bibitem{Nister2004f}
D.~Nist{\'{e}}r, O.~Naroditsky, and J.~Bergen, ``{Visual Odometry},'' in
  \emph{IEEE CVPR}, 2004, pp. 652--659.

\bibitem{Forster2014}
C.~Forster, M.~Pizzoli, and D.~Scaramuzza, ``{SVO: Fast Semi-Direct Monocular
  Visual Odometry},'' in \emph{IEEE ICRA}, 2014, pp. 15--22.

\bibitem{Mur-Artal2015b}
R.~Mur-Artal, J.~M.~M. Montiel, and J.~D. Tardos, ``{ORB-SLAM: A Versatile and
  Accurate Monocular SLAM System},'' \emph{IEEE T-RO}, vol.~31, no.~5, pp.
  1147--1163, 2015.

\bibitem{Wang2017}
S.~Wang, R.~Clark, H.~Wen, and N.~Trigoni, ``{DeepVO: Towards End-to-End Visual
  Odometry with Deep Recurrent Convolutional Neural Networks},'' in \emph{IEEE
  ICRA}, 2017.

\bibitem{saputra19clvo}
M.~R.~U. Saputra, P.~P. B.~d. Gusmao, S.~Wang, A.~Markham, and N.~Trigoni,
  ``Learning monocular visual odometry through geometry-aware curriculum
  learning,'' in \emph{IEEE ICRA}, 2019.

\bibitem{almalioglu2018ganvo}
Y.~Almalioglu, M.~R.~U. Saputra, P.~P. de~Gusmao, A.~Markham, and N.~Trigoni,
  ``Ganvo: Unsupervised deep monocular visual odometry and depth estimation
  with generative adversarial networks,'' in \emph{IEEE ICRA}, 2019.

\bibitem{zhan2018unsupervised}
H.~Zhan, R.~Garg, C.~Saroj~Weerasekera, K.~Li, H.~Agarwal, and I.~Reid,
  ``Unsupervised learning of monocular depth estimation and visual odometry
  with deep feature reconstruction,'' in \emph{IEEE CVPR}, 2018, pp. 340--349.

\bibitem{kanellakis2016evaluation}
C.~Kanellakis and G.~Nikolakopoulos, ``Evaluation of visual localization
  systems in underground mining,'' in \emph{IEEE MED}, 2016, pp. 539--544.

\bibitem{peynot2009towards}
T.~Peynot, J.~Underwood, and S.~Scheding, ``Towards reliable perception for
  unmanned ground vehicles in challenging conditions,'' in \emph{IEEE/RSJ
  IROS}, 2009, pp. 1170--1176.

\bibitem{quater2014light}
P.~B. Quater, F.~Grimaccia, S.~Leva, M.~Mussetta, and M.~Aghaei, ``Light
  unmanned aerial vehicles (uavs) for cooperative inspection of pv plants,''
  \emph{IEEE J-PV}, vol.~4, no.~4, pp. 1107--1113, 2014.

\bibitem{yu2013camera}
Z.~Yu, S.~Lincheng, Z.~Dianle, Z.~Daibing, and Y.~Chengping, ``Camera
  calibration of thermal-infrared stereo vision system,'' in \emph{ISDA}.\hskip
  1em plus 0.5em minus 0.4em\relax IEEE, 2013, pp. 197--201.

\bibitem{averbuch2007scene}
A.~Averbuch, G.~Liron, and B.~Z. Bobrovsky, ``Scene based non-uniformity
  correction in thermal images using kalman filter,'' \emph{Image and Vision
  Computing}, vol.~25, no.~6, pp. 833--851, 2007.

\bibitem{saputra2018visual}
M.~R.~U. Saputra, A.~Markham, and N.~Trigoni, ``Visual slam and structure from
  motion in dynamic environments: A survey,'' \emph{ACM CSUR}, vol.~51, no.~2,
  p.~37, 2018.

\bibitem{chen2019selective}
C.~Chen, S.~Rosa, Y.~Miao, C.~X. Lu, W.~Wu, A.~Markham, and N.~Trigoni,
  ``Selective sensor fusion for neural visual-inertial odometry,'' in
  \emph{IEEE CVPR}, 2019, pp. 10\,542--10\,551.

\bibitem{mouats2015thermal}
T.~Mouats, N.~Aouf, L.~Chermak, and M.~A. Richardson, ``Thermal stereo odometry
  for uavs,'' \emph{IEEE Sensors Journal}, vol.~15, no.~11, pp. 6335--6347,
  2015.

\bibitem{khattak2019keyframe}
S.~Khattak, C.~Papachristos, and K.~Alexis, ``Keyframe-based direct
  thermal-inertial odometry,'' in \emph{IEEE ICRA}, 2019.

\bibitem{borges2016practical}
P.~V.~K. Borges and S.~Vidas, ``Practical infrared visual odometry,''
  \emph{IEEE T-ITS}, vol.~17, no.~8, pp. 2205--2213, 2016.

\bibitem{poujol2016visible}
J.~Poujol, C.~A. Aguilera, E.~Danos, B.~X. Vintimilla, R.~Toledo, and A.~D.
  Sappa, ``A visible-thermal fusion based monocular visual odometry,'' in
  \emph{Robot 2015: Second Iberian Robotics Conference}.\hskip 1em plus 0.5em
  minus 0.4em\relax Springer, 2016, pp. 517--528.

\bibitem{khattak2019visual}
S.~Khattak, C.~Papachristos, and K.~Alexis, ``Visual-thermal landmarks and
  inertial fusion for navigation in degraded visual environments,'' in
  \emph{2019 IEEE Aerospace Conference}.\hskip 1em plus 0.5em minus 0.4em\relax
  IEEE, 2019, pp. 1--9.

\bibitem{papachristos2018thermal}
C.~Papachristos, F.~Mascarich, and K.~Alexis, ``Thermal-inertial localization
  for autonomous navigation of aerial robots through obscurants,'' in
  \emph{ICUAS}.\hskip 1em plus 0.5em minus 0.4em\relax IEEE, 2018, pp.
  394--399.

\bibitem{valada2018deep}
A.~Valada, N.~Radwan, and W.~Burgard, ``Deep auxiliary learning for visual
  localization and odometry,'' in \emph{IEEE ICRA}.\hskip 1em plus 0.5em minus
  0.4em\relax IEEE, 2018, pp. 6939--6946.

\bibitem{radwan2018vlocnet}
N.~Radwan, A.~Valada, and W.~Burgard, ``Vlocnet++: Deep multitask learning for
  semantic visual localization and odometry,'' \emph{IEEE RAL}, vol.~3, no.~4,
  pp. 4407--4414, 2018.

\bibitem{clark2017vinet}
R.~Clark, S.~Wang, H.~Wen, A.~Markham, and N.~Trigoni, ``Vinet: Visual-inertial
  odometry as a sequence-to-sequence learning problem,'' in \emph{AAAI}, 2017.

\bibitem{Zhou2017}
T.~Zhou, M.~Brown, N.~Snavely, and D.~G. Lowe, ``{Unsupervised Learning of
  Depth and Ego-Motion from Video},'' in \emph{IEEE CVPR}, 2017.

\bibitem{li2018undeepvo}
R.~Li, S.~Wang, Z.~Long, and D.~Gu, ``Undeepvo: Monocular visual odometry
  through unsupervised deep learning,'' in \emph{IEEE ICRA}.\hskip 1em plus
  0.5em minus 0.4em\relax IEEE, 2018, pp. 7286--7291.

\bibitem{yang2018deep}
N.~Yang, R.~Wang, J.~Stuckler, and D.~Cremers, ``Deep virtual stereo odometry:
  Leveraging deep depth prediction for monocular direct sparse odometry,'' in
  \emph{ECCV}, 2018, pp. 817--833.

\bibitem{feng2019sganvo}
T.~Feng and D.~Gu, ``Sganvo: Unsupervised deep visual odometry and depth
  estimation with stacked generative adversarial networks,'' \emph{arXiv
  preprint arXiv:1906.08889}, 2019.

\bibitem{hoffman2016learning}
J.~Hoffman, S.~Gupta, and T.~Darrell, ``Learning with side information through
  modality hallucination,'' in \emph{IEEE CVPR}, 2016, pp. 826--834.

\bibitem{choi2017learning}
C.~Choi, S.~Kim, and K.~Ramani, ``Learning hand articulations by hallucinating
  heat distribution,'' in \emph{IEEE ICCV}, 2017, pp. 3104--3113.

\bibitem{lezama2017not}
J.~Lezama, Q.~Qiu, and G.~Sapiro, ``Not afraid of the dark: Nir-vis face
  recognition via cross-spectral hallucination and low-rank embedding,'' in
  \emph{IEEE CVPR}, 2017, pp. 6628--6637.

\bibitem{connor1994recurrent}
J.~T. Connor, R.~D. Martin, and L.~E. Atlas, ``Recurrent neural networks and
  robust time series prediction,'' \emph{IEEE T-NN}, vol.~5, no.~2, pp.
  240--254, 1994.

\bibitem{Dosovitskiy2016}
A.~Dosovitskiy, P.~Fischery, E.~Ilg, P.~Hausser, C.~Hazirbas, V.~Golkov,
  P.~V.~D. Smagt, D.~Cremers, and T.~Brox, ``{FlowNet: Learning Optical Flow
  with Convolutional Networks},'' in \emph{IEEE ICCV}, vol. 11-18-Dece, 2016,
  pp. 2758--2766.

\bibitem{deeppco2019}
W.~Wang, M.~R.~U. Saputra, P.~Zhao, P.~Gusmao, B.~Yang, C.~Chen, A.~Markham,
  and N.~Trigoni, ``Deeppco: End-to-end point cloud odometry through deep
  parallel neural network,'' \emph{IEEE/RSJ IROS}, 2019.

\bibitem{srivastava2014dropout}
N.~Srivastava, G.~Hinton, A.~Krizhevsky, I.~Sutskever, and R.~Salakhutdinov,
  ``Dropout: A simple way to prevent neural networks from overfitting,''
  \emph{JMLR}, vol.~15, no.~1, pp. 1929--1958, 2014.

\bibitem{saputra2019distilling}
M.~R.~U. Saputra, P.~P. de~Gusmao, Y.~Almalioglu, A.~Markham, and N.~Trigoni,
  ``Distilling knowledge from a deep pose regressor network,'' \emph{arXiv
  preprint arXiv:1908.00858}, 2019.

\bibitem{huber1992robust}
P.~J. Huber, ``Robust estimation of a location parameter,'' in
  \emph{Breakthroughs in statistics}.\hskip 1em plus 0.5em minus 0.4em\relax
  Springer, 1992, pp. 492--518.

\bibitem{qin2018vins}
T.~Qin, P.~Li, and S.~Shen, ``Vins-mono: A robust and versatile monocular
  visual-inertial state estimator,'' \emph{IEEE T-RO}, vol.~34, no.~4, pp.
  1004--1020, 2018.

\bibitem{sturm2012benchmark}
J.~Sturm, N.~Engelhard, F.~Endres, W.~Burgard, and D.~Cremers, ``A benchmark
  for the evaluation of rgb-d slam systems,'' in \emph{IEEE/RSJ IROS}, 2012,
  pp. 573--580.

\bibitem{chen2018ionet}
C.~Chen, X.~Lu, A.~Markham, and N.~Trigoni, ``Ionet: Learning to cure the curse
  of drift in inertial odometry,'' in \emph{AAAI}, 2018.

\bibitem{bloesch2015robust}
M.~Bloesch, S.~Omari, M.~Hutter, and R.~Siegwart, ``Robust visual inertial
  odometry using a direct ekf-based approach,'' in \emph{IEEE/RSJ IROS}, 2015,
  pp. 298--304.

\bibitem{bijelic2019seeing}
M.~Bijelic, F.~Mannan, T.~Gruber, W.~Ritter, K.~Dietmayer, and F.~Heide,
  ``Seeing through fog without seeing fog: Deep sensor fusion in the absence of
  labeled training data,'' \emph{arXiv preprint arXiv:1902.08913}, 2019.

\bibitem{wahlstrom2019zero}
J.~Wahlstr\"om, I.~Skog, F.~Gustafsson, A.~Markham, and N.~Trigoni,
  ``Zero-velocity detection -- {A B}ayesian approach to adaptive
  thresholding,'' \emph{IEEE Sensors Letters}, vol.~3, no.~6, 2019.

\end{thebibliography}
\bibliographystyle{IEEEtran}

%








\end{document}